\documentclass[lettersize,journal]{IEEEtran}
\usepackage{amsmath,amsfonts}
\usepackage{array}
\usepackage{textcomp}
\usepackage{stfloats}
\usepackage{url}
\usepackage{verbatim}
\usepackage{graphicx}
\usepackage{booktabs} 
\usepackage{array}      
\usepackage{siunitx}     
\usepackage{cite}
\usepackage{subfigure}
\usepackage{epsfig}
\usepackage{amssymb}
\usepackage[ruled,linesnumbered]{algorithm2e}
\usepackage{algpseudocode}  
\usepackage{bm}
\usepackage{multirow}
\usepackage{color}
\usepackage{hyperref}

\begin{document}

\title{Fine-Grained Behavior and Lane Constraints Guided Trajectory Prediction Method}

\author{Wenyi Xiong$^{1}$, Jian Chen$^{2}$, \IEEEmembership{Senior Member, IEEE}, {Ziheng Qi}$^{3}$ 
\thanks{This work is partially supported by Shenzhen Science and Technology Program under Grant No. JCYJ20250604144718024 and No. KQTD20221101093557010, and the Guangdong Provincial Key Laboratory of Fully Actuated System Control Theory and Technology under Grant 2024B1212010002.}
\thanks{$^{1}$Wenyi Xiong is with the School of Mechanical Engineering, Zhejiang University, Hangzhou 310027, China. }
\thanks{$^{2}$Jian Chen is with the Guangdong Provincial Key Laboratory of Fully
Actuated System Control Theory and Technology, School of Automation
and Intelligent Manufacturing, Southern University of Science and
Technology, Shenzhen 518055, China. He is also with the School of
Mechanical Engineering, Zhejiang University, Hangzhou 310058, China.
(e-mail: chenj8@sustech.edu.cn). } 
\thanks{$^{3}$Ziheng Qi is with Leapmotor Technology, Hangzhou 310052, China.}

}


\maketitle

\begin{abstract}
Trajectory prediction, as a critical component of autonomous driving systems, has attracted considerable attention in recent years. Existing prediction methods typically focus on extracting richer scene representations or introducing trajectory-level guidance through destinations, goals, or motion modes. However, future vehicle motion is continuously influenced by both evolving driving intentions and structured lane constraints. Such dynamic factors are difficult to characterize using coarse trajectory-level representations alone, which may limit prediction accuracy.
To address this challenge, we propose BLNet, a novel fine-grained trajectory prediction framework that explicitly models behavioral evolution and lane topology constraints throughout the prediction horizon. Specifically, a dual-stream architecture is designed to generate timestamp-level behavior-state queries and lane queries, enabling future motion to be characterized from both behavioral and lane-constrained perspectives. The generated queries are supervised by dedicated auxiliary objectives and jointly guide multimodal trajectory generation. Furthermore, a two-stage decoder is introduced to first produce trajectory proposals and then perform point-level refinement by incorporating lane continuity and future motion features.
Extensive experiments on the nuScenes and Argoverse benchmarks demonstrate the effectiveness of the proposed framework. The results show that BLNet achieves a favorable balance between prediction accuracy and inference efficiency while maintaining competitive forecasting performance.

\end{abstract}


\section{Introduction}
\IEEEPARstart{T}{he} prediction task has received significant attention from researchers 
\cite{chen2022vehicle,guo2022vehicle,liu2024laformer} as an important intermediate step 
between the perception and decision-making processes.  However, the complexity of traffic scenarios
 (e.g. complex lane constraints and traffic rules) and the uncertainty of the future trajectory modes of vehicles makes forecasting still a major challenge.  
Today, great success has been achieved in the field of trajectory prediction by utilizing rapidly 
evolving deep learning techniques.  
Some frameworks such as Graph Neural Networks (GNN) \cite{li2019grip}, Convolutional Neural 
Networks (CNN) \cite{xiong2023hierarchical}, etc. are applied to mine the interactions between 
traffic elements. In addition, the attention mechanism \cite{zhou2022hivt} is gradually being widely
 used by predictive models. Due to its effective modeling of interaction features, the prediction 
 performance is further improved.
\begin{figure}[h]
	\centering
	\includegraphics[width=9cm]{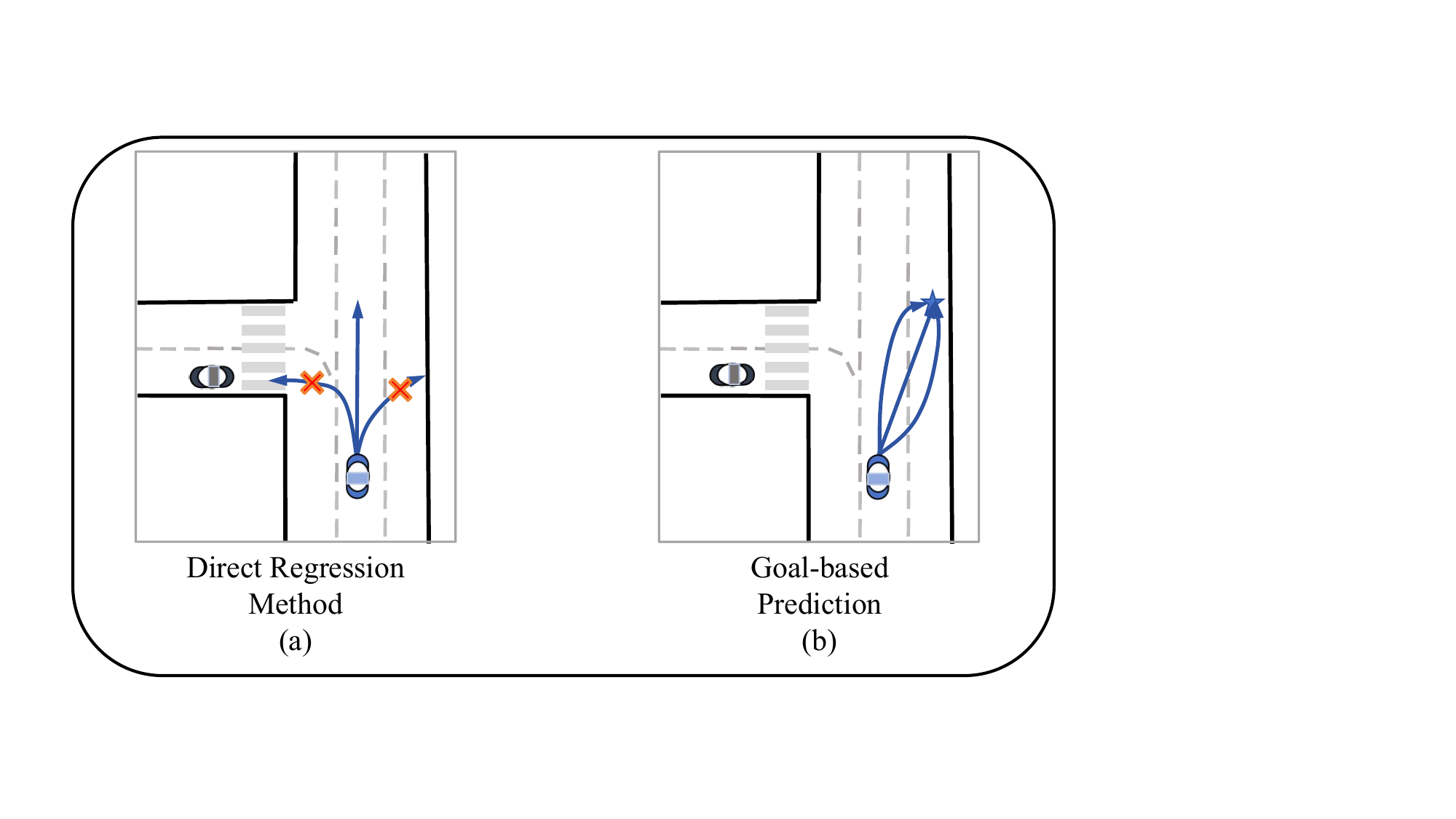}
	\caption{Illustration of different prediction methods. 
	(a) Direct regression. Unreasonable trajectories may be generated due to 
	the lack of a spatial prior. (b) Goal-based method. There can be different
	paths to the same goal. } 
	\label{fig:methods}
\end{figure}

Most trajectory prediction networks fall into two categories: direct regression and goal-based methods.
Direct regression prediction algorithms \cite{gao2020vectornet,song2020pip} focus on extracting
 multi-agents interactions and complex scene constraints and directly regress to obtain predicted
  trajectories  as shown in Fig. \ref{fig:methods} (a). For example, HiVT \cite{zhou2022hivt} efficiently
   models the features of a large number of agents in a scene by local context extraction and global
    interaction modeling, respectively.  A reusable multi-context gating fusion module  is developed
	 by Multipath++ \cite{varadarajan2022multipath++} to extract interaction features and environment
	  features in a scene. Although much progress has been made, without the guidance of a spatial prior,
	   it is difficult for such algorithms to accurately predict the multimodality of future trajectories
	    through a single feature-aggregated query, and some implausible future trajectories are thus 
		generated.
 In order to solve the problem of modal uncertainty, many researchers begin to consider
  providing  priori destinations for prediction, and thus goal-based prediction 
  algorithms emerge as shown in Fig. \ref{fig:methods} (b). 
  Such algorithms \cite{zhao2021tnt,gu2021densetnt,aydemir2023adapt} guide the 
  network to generate a modal-determined trajectory by predicting the possible future
   destinations or target lanes. The destinations can be manually 
   produced or dynamically predicted by the network. For example, 
   DenseTNT \cite{gu2021densetnt} manually produces a large number of target points 
   scattered in the map. However, the performance of prediction depends greatly on 
   the number and quality of candidates.
GOHOME \cite{gilles2022gohome} utilizes high-definition map graphics and sparse
 mapping to generate heatmap outputs to predict future target points for final 
 prediction.
 Ltp\cite{wang2022ltp} predicts the likelihood of each lane segment and from this,
  trajectories converging to the most probable lane segments are generated and 
  selected. However, the future motion of a vehicle is dynamically changing and is
   driven by both its own intentions and the structure of the roadway. This leads to
    the fact that there can be different paths leading to the same goal. Trajectories
	 generated solely from a single destination or a single lane segment tend to 
	 ignore detailed spatiotemporal interactions and lane constraints, which degrades
	  model performance.

\begin{figure}
	\centering
	\includegraphics[width=9cm]{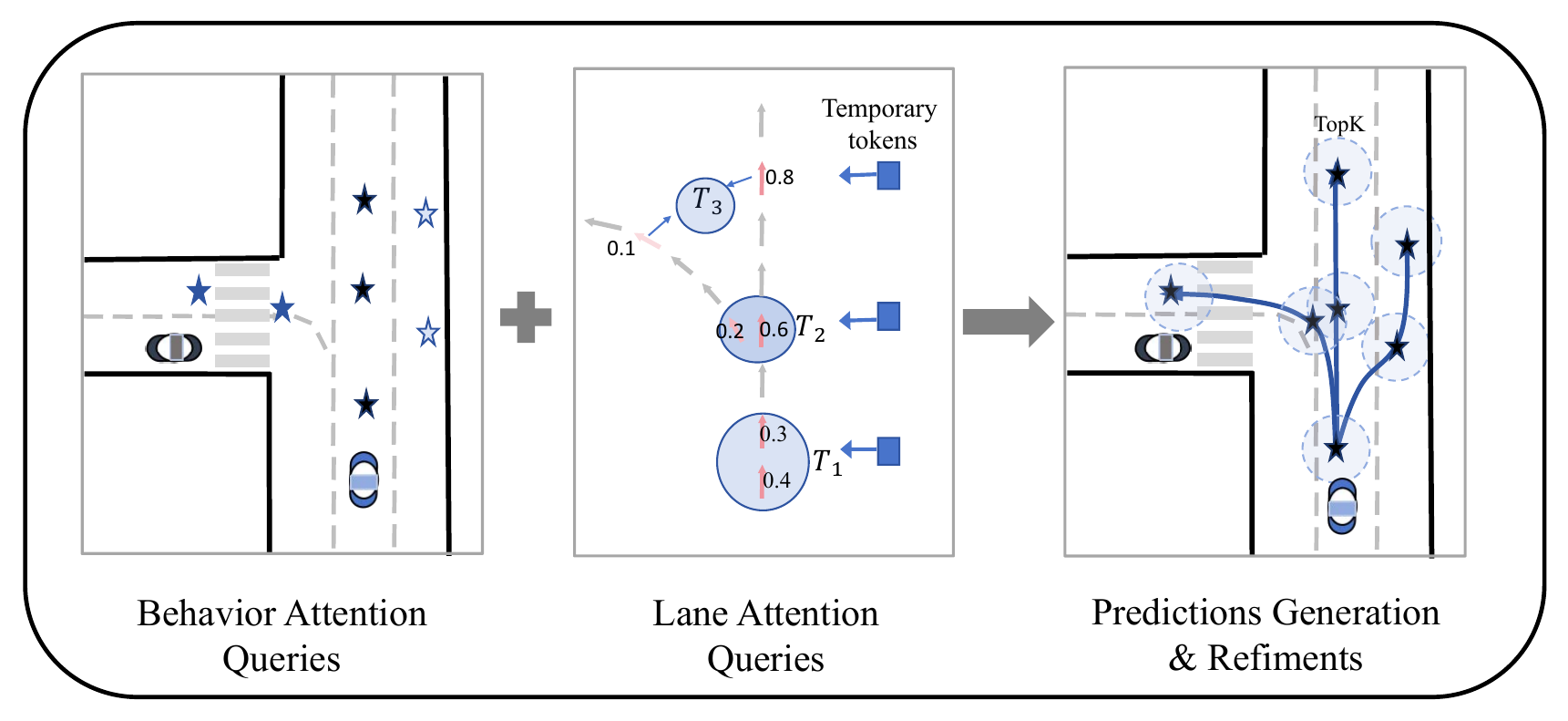}
	\caption{An overview of the proposed algorithm framework. 
	Unlike conventional query-based prediction methods that
mainly utilize trajectory-level guidance, the proposed framework
employs fine-grained behavior-state queries and lane queries
to characterize future motion throughout the prediction horizon. Finally, the continuity of the lane and 
	  future motion features are aggregated to refine the trajectories at the point-level.} 
	\label{fig:vis}
\end{figure}

In order to solve the problems of the two types of methods mentioned above, 
we propose a novel attention based network as shown in Fig. \ref{fig:vis}.
 Inspired by objective factors affecting future trajectories, 
 we simultaneously aggregate detailed lane information and driver intent 
 information to predict dynamically evolving target future motions. 
 Specifically, two attention branches are proposed to generate 
 fine-grained behavior state queries and lane queries. 
 The behavior state query reflects the motion intent of 
 the target vehicle at each timestamp under the supervision of 
 the ground truth location. Lane queries aggregate structured lane 
 constraints under the supervision of the lanes that the vehicle will 
 reach in each future timestamp. 
 The two queries are aggregated and fed into a two-stage decoder.
 The two-stage decoder first predicts future trajectories and further refines 
 the predictions at the point level to make them more reasonable by exploiting the
  future motion features and the continuity of lane constraints.

Based on the above discussion, our contribution is summarized below:

\begin{itemize}
\item We propose a fine-grained trajectory prediction formulation that jointly models behavioral evolution and lane topology constraints throughout the prediction horizon. Specifically, timestamp-level behavior-state queries and lane queries are introduced to explicitly characterize future motion states, enabling trajectory prediction to be guided by both evolving driving intentions and structured lane constraints.

	\item A new dual-stream attention network is proposed to acquire behavior 
	state queries and lane queries. Under the supervision of two auxiliary loss functions,
	 these queries are aligned to the vehicle future behaviors and the lane constraints
	  at each timestamp, respectively.
	
	\item In the decoder, the future motion and the continuity of lane 
	constraints are further incorporated into the network to achieve point-level
	 trajectory refinement.
	
	\item Extensive experimental and ablation studies on the NuScenes and Argoverse benchmarks demonstrate the effectiveness of the proposed framework. The results show that BLNet achieves a favorable balance between prediction accuracy and inference efficiency while maintaining strong trajectory forecasting performance.

\end{itemize}

\section{Related Works}

\subsection{Scene Feature Modeling}
In order to obtain accurate predicted trajectories, it is usually necessary to have a 
fine-grained modeling of the traffic scenario.
CNNs are widely used due to their superior ability to extract localized features.
SoPhie \cite{sadeghian2019sophie} utilizes CNN to extract information from scene images.
Multi-agent future interaction relationships and future planning information of 
controllable vehicles are modeled in PiP \cite{song2020pip} using CNN.
However, CNNs are limited by their convolutional blocks and ignore the detailed interaction 
relationships.
In order to solve this problem, researchers try to utilize GCNs to obtain a relationship modeling of
 the nodes in the scene.
Grip \cite{li2019grip} and Grip++ \cite{li2019grip++} introduce a static graph and a dynamic graph
 structure to describe the interactions between transportation agents, respectively.
Xu et al. \cite{xu2022adaptive} propose a transferable graph neural network that jointly
 performs motion forecasting and domain alignment.
In recent years, transformers have begun to be introduced into the field of
 trajectory prediction to cope with spatiotemporal information processing.
 \cite{zhang2024simpl,aydemir2023adapt}.
HiVT \cite{zhou2022hivt} divides the problem into attention-based local context 
extraction and global feature interaction to achieve efficient trajectory prediction 
results.
HPNet \cite{tang2024hpnet} models overlapping history frames using the 
attention mechanism to achieve a multi-frame prediction with continuity.
In order to extract more interactions, the common transformer 
is further extended, e.g., by introducing relative position coding, combining the
 transformer with graph convolution, 
 etc. \cite{zhang2024simpl,shi2020multimodal,yang2023long}.
For example, Zhou et al. \cite{zhou2024edge} proposed a new edge-enhanced
 interaction module to model the relative position information between nodes
  in the scene.
In this paper, we construct dual-stream attention branches to generate behavior state
 queries and lane queries at each future timestamp in parallel to guide trajectory
  prediction.

\subsection{Direct Regression Prediction Method}
This kind of method \cite{zhang2024simpl,yang2023long,mohamed2020social} mines
 the interaction features among agents and directly regresses to obtain the 
 final predicted trajectory, as shown in Fig. \ref{fig:methods} (a). 
  Many attempts are made to mine more detailed interaction features. 
  For example, Trajectron++ \cite{salzmann2020trajectron++} uses a modular 
  graph-structured recursive model to model dynamic interactions in the scene.
 Hdgt \cite{jia2023hdgt} achieves more accurate predictions 
 by modeling the scenario as a heterogeneous graph with different nodes and edges. 
 Simpl \cite{zhang2024simpl} designs a symmetric fusion transformer for efficient
  viewpoint-invariant scene modeling.
Zhang  et al. \cite{zhang2024edge} explicitly model the edge information between 
traffic participants, making the network aware of the proximity relationship between
 them.
However, due to the lack of spatial  priors, these algorithms converge slowly when 
faced with multi-modal trajectory prediction. In addition, uncertainty in driving 
intentions can affect the predictive effectiveness of such algorithms, which in turn
 can produce some unreasonable predictions. Our algorithm reduces the impact of 
 uncertainty by combining fine-grained behavior state and lane queries to guide predictions.
 
 \subsection{Goal-Based Prediction Method}
 
 This type of approach reduces the complexity of the prediction
  problem by decomposing the complex trajectory prediction problem into goal 
  regression and trajectory regression, as shown in Fig. \ref{fig:methods} (b).
   In terms of the types of regression destinations, goal-based methods can be categorized into
    goal-point-based prediction and goal-lane-based prediction.

As pioneers of the goal-point-based approaches,
TNT \cite{zhao2021tnt} and DenseTNT \cite{gu2021densetnt} produce a large number of manual
 candidate target points. A function is first utilized to predict the probability of each point,
  and a number of the most likely points are selected to perform the trajectory completion. 
In order to get rid of a large number of redundant manual goal points for prediction efficiency, 
algorithms based on adaptive goal points have been proposed.
MTR \cite{shi2022motion} proposes learnable spatial prior queries for simultaneous intent recognition
 and trajectory refinement.
Adapt \cite{aydemir2023adapt} enhances the accuracy of the endpoint-based prediction algorithm by a 
gradient-stopping training strategy.
The prediction accuracy of such methods depends largely on the quality of the target point. For example,
 an unreasonable target point can lead to the generation of a trajectory that violates the lane 
 constraints.

Vehicle trajectories on structured lanes are significantly affected by path constraints. Therefore, based on the guidance of a goal lane \cite{wang2022ltp},
 we can get more reasonable predictions. Inspired by this, PGP \cite{deo2022multimodal} explores the trajectory of a vehicle from one lane node to another by training
  a strategy. Yet such an exploration also comes with a significant computational cost. GOHOME \cite{gilles2022gohome} obtains a heat map of the target lanes to guide trajectory prediction and reduce computational cost. Li et al. \cite{li2024efficient} propose a Dual-Stream Cross Attention to obtain the most likely K target lanes and concatenate them with the target encoding to predict future motion. However, lanes are highly structured and often do not fully reflect fine-grained driver intent, which in turn produces sub-optimal results.

In summary, while goal-based approaches can reduce forecasting uncertainty to some extent, the fact that the future trajectory of a goal is dynamically evolving and is often influenced by both its own intentions and structured lane constraints creates a challenge for forecasting. Thus, our method combines both fine-grained lane constraints and driver behavioral intentions.

\section{Proposed Model}

\begin{figure*}
	\centering
	\includegraphics[width=16cm]{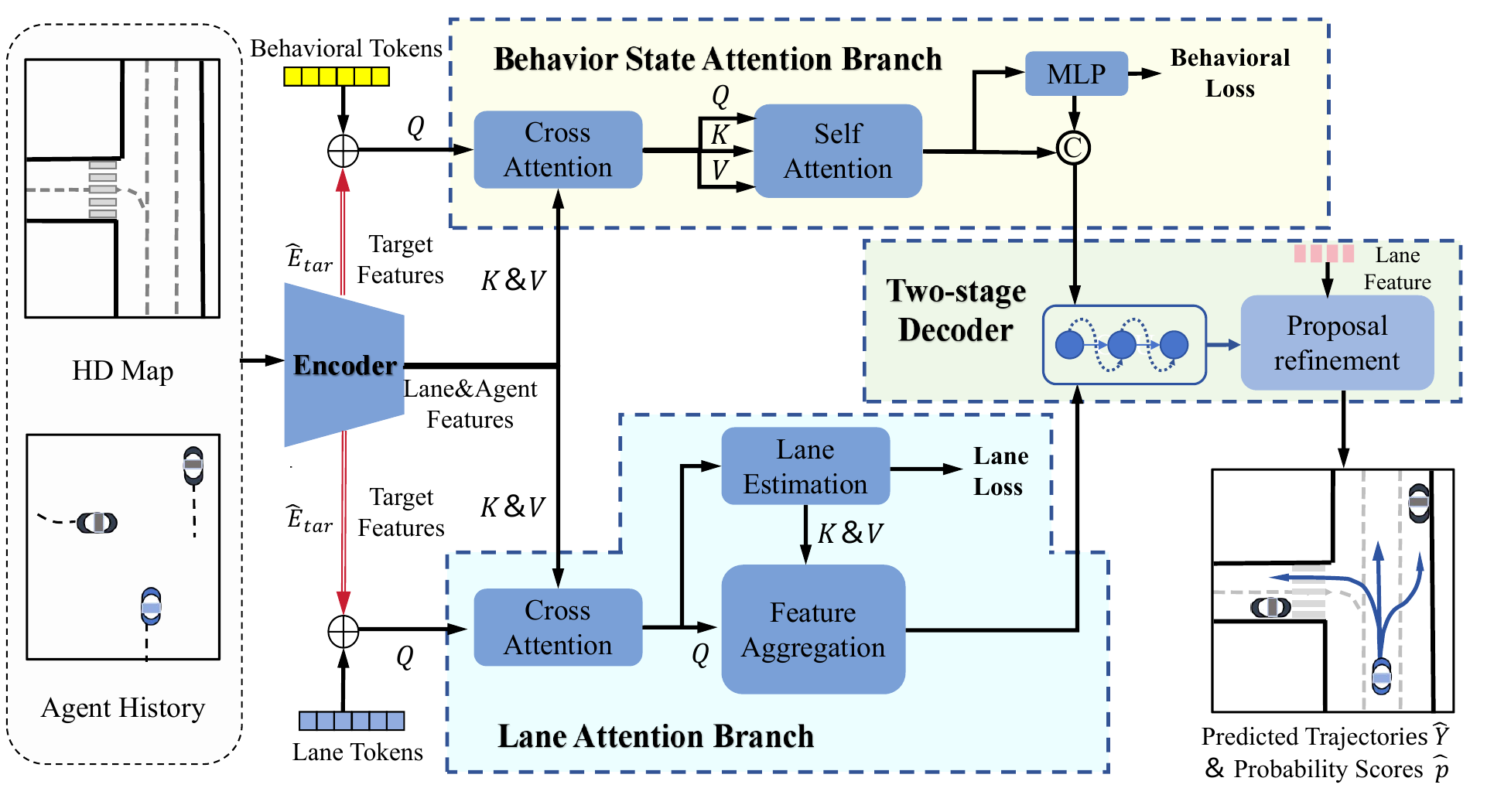}
	\caption{The pipeline of our proposed algorithm. After vectorization,
	  agent and map information are fed into the encoder and processed using RNN 
	  and transformer to get its corresponding high dimensional feature information.
	   Subsequently, we designed two attention branches (lane attention branch and
	    behavior state attention branch) to obtain temporal lane constraints and behavior
		 queries. Finally, a two-stage decoder is designed to predict the target trajectories and refine them at the point-level
		 using the continuity of the lane and future behavior.} 
	\label{fig:frame}
\end{figure*}

We propose a novel network framework that utilizes 
fine-grained behavioral and lane constraints to guide the 
 trajectory predictions as shown in Fig. \ref{fig:frame}.
 In the following, we will first introduce the problem setup. 
 Then, we will elaborate on the four components of the network in turn: 
 the encoder, the behavior state attention branch, the lane attention branch, 
 and the two-stage trajectory decoder. 
 Finally, the model training details are introduced.

\subsection{Problem Setup}

Given the HD map information and historical observations of surrounding traffic participants, the objective of trajectory prediction is to estimate the future motion of a target vehicle over a prediction horizon of $T_f$ timesteps. Specifically, the future trajectory is represented as

\begin{equation}
\mathcal{Y} \in \mathbb{R}^{T_f\times 2},
\end{equation}

where $T_f$ denotes the prediction horizon and each trajectory point contains the two-dimensional spatial coordinates of the target vehicle at the corresponding future timestamp. Based on the observed scene context, the proposed framework aims to generate accurate and feasible future trajectory predictions for the target agent.

\subsection{Traffic Scene Vectorization and Encoding}

\begin{algorithm}[h]
\caption{Preprocessing Pipeline for Lane and Agent Vectorization}
	\label{alg:vec}
\SetKwFunction{FVec}{VectorizePipeline}
\SetKwProg{Fn}{function}{:}{}
\textbf{Input}: HD lane polylines, agent historical motion sequences
\textbf{Hyperparams}: $S$, $C_m$, $C_v$, $T_h$
\textbf{Output}: Map tensor $\mathcal{M}\in\mathbb{R}^{N_m\times S\times C_m}$, Agent tensor $\mathcal{H}\in\mathbb{R}^{N_v\times T_h\times C_v}$

\Fn{\FVec{HD\_lanes, Agent\_seqs, $S$, $C_m$, $C_v$, $T_h$}}{
    \BlankLine
    \% Lane vectorization
    $L_{seg} \leftarrow \emptyset$ \\
    \For{each lane polyline $\in$ HD\_lanes}{
        Densely subdivide polyline into dense coordinate points, then split into local lane segments via sliding window with length $S$ \\
        \For{each segment $seg$}{
            Extract $C_m$-dimensional geometric and semantic features for each point in the segment, and pad short segments to $S$ points to obtain a $S\times C_m$ feature matrix \\
            Append the segment feature matrix to $L_{seg}$
        }
    }
    $\mathcal{M} \leftarrow \mathtt{Stack}(L_{seg})$
    \BlankLine
    \% Agent motion vectorization
    $A_{hist} \leftarrow \emptyset$ \\
    \For{each agent $v \in$ Agent\_seqs}{
        Extract $C_v$-dimensional motion and attribute features over $T_h$ historical timestamps to construct a temporal feature sequence \\
        Append the sequence to $A_{hist}$
    }
    $\mathcal{H} \leftarrow \mathtt{Stack}(A_{hist})$
    \BlankLine
    \Return $\mathcal{M},\ \mathcal{H}$
}
\end{algorithm}

\begin{figure}
	\centering
	\includegraphics[width=8cm]{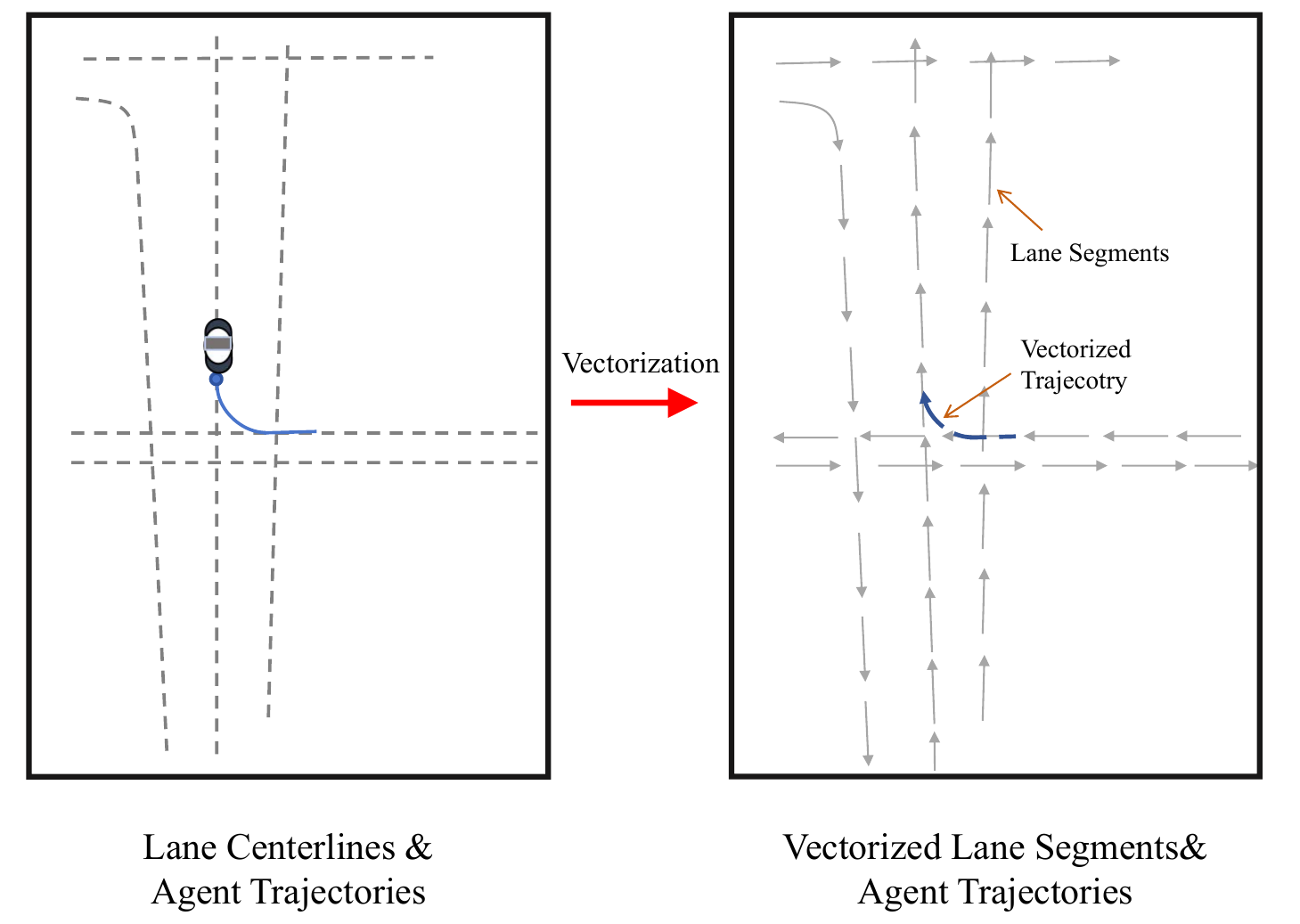}
	\caption{Scene information vectorization. After vectorization, lane segments and trajectories are represented by a vector.} 
	\label{fig:vec}
\end{figure}

Inspired by Vectornet \cite{gao2020vectornet}, 
we vectorize scene elements as shown in Fig. \ref{fig:vec}  to facilitate prediction networks to encode them 
to obtain high-dimensional features.

\textbf{Map Vectorization:}
Specifically, lane centerlines are first subdivided into dense points 
and then partitioned into multiple local lane segments. Each local 
segment is subsequently converted into vector representations, yielding
$\mathcal{M}\in\mathbb{R}^{N_m\times S\times C_m}$,
where $N_m$ denotes the number of lane segments, $S$ denotes the 
number of coordinate points within each segment, and $C_m$ denotes
 the feature dimension. Each lane vector stores both geometric coordinates and semantic attributes including turn direction, intersection marker and traffic control flag, and records the connection relation with previous lane segments; connected vectors jointly describe the complete topology of the lane centerline. As this process does not rely on predefined road templates, straight roads, curved lanes, intersections, merges, and roundabouts can all be represented within the same vectorized framework.

\textbf{Agents Vectorization:}
Similar to the processing of map information, historical observations of traffic participants are vectorized and formulated as
$\mathcal{H}\in\mathbb{R}^{N_v\times T_h\times C_v}$,
where $N_v$ denotes the number of traffic participants, $T_h$ denotes the historical observation time domain, and $C_v$ denotes the feature dimension. For each timestamp, the vector stores the coordinates of the start and end points of the motion segment together with agent attribute labels, which jointly characterize the local motion evolution of each target between consecutive observations and encode temporal motion patterns. Such a representation remains applicable across diverse traffic layouts, since it is defined purely based on real observed motion rather than preset road-shape assumptions.
The full vector generation preprocessing procedure is detailed in Algorithm \ref{alg:vec} in an algorithm-style manner.

The vectorized map representation $\mathcal{M}$ and agent representation $\mathcal{H}$ are first encoded into latent feature embeddings, yielding map features $E_{\mathcal{M}}\subseteq \mathbb{R}^{N_m\times C_e}$ and agent features $E_{\mathcal{H}}\subseteq \mathbb{R}^{N_v\times C_e}$, where $C_e$ denotes the hidden feature dimension.
Inspired by LaneGCN \cite{liang2020learning}, we further model the interactions between lane segments and traffic participants through a bidirectional attention-based fusion architecture. Specifically, map features are treated as queries, while agent features serve as keys and values, yielding the interaction-enhanced map representations:
\begin{equation}
\hat{E}_\mathcal{M}
=
\mathcal{I}
\left(
Q=E_\mathcal{M},
K=V=E_\mathcal{H}
\right)
\end{equation}
Through this cross-attention operation, each lane representation is enriched with relevant agent motion information. Conversely, the interaction-enhanced map representations are further aggregated into agent features to obtain the interactive agent encodings:
\begin{equation}
\hat{E}_\mathcal{H}
=
\mathcal{I}
\left(
Q=E_\mathcal{H},
K=V=\hat{E}_\mathcal{M}
\right)
\end{equation}
The encoding corresponding to the target vehicle is denoted as $\hat{E}_{tar}$. Finally, the interaction-enhanced map and agent representations are combined to form the overall scene encoding:
\begin{equation}
\hat{E}
=
\left\{
\hat{E}_{\mathcal{M}},
\hat{E}_{\mathcal{H}}
\right\}.
\end{equation}

 \subsection{Behavior State Attention Branch}
For traffic participants, driving intention is one of the key
factors influencing future motion evolution \cite{zhang2024decoupling}.
Previous goal-point-based approaches \cite{gu2021densetnt,aydemir2023adapt}
reduce trajectory uncertainty by providing destination priors.
However, future trajectories are inherently dynamic and continuously evolving,
and different motion patterns may still emerge under the same destination.
Therefore, relying solely on goal-level guidance may be insufficient to
characterize the detailed evolution of future motion.

To address this issue, we propose a behavior state attention branch
that provides fine-grained behavioral priors throughout the prediction horizon.
Instead of directly representing future trajectories or destinations, similar to \cite{carion2020end},
the proposed learnable behavior tokens  are designed to capture latent
behavioral evolution patterns that may influence future motion generation.
It should be noted that these tokens do not correspond to predefined
semantic behaviors, but are
automatically learned from data through end-to-end optimization.
Considering the multi-modal nature of future trajectories, we assign
$K$ learnable tokens $B_{tokens}$ to the target encoding to obtain
the initial behavior queries $\mathcal{Q}_{b}$:
\begin{equation}
\mathcal{Q}_b = \hat{E}_{tar}+B_{tokens}
\end{equation}

 We use scene encoding $\hat{E}$ as key and values to perform a cross attention calculation with $\mathcal{Q}_{b}$ to obtain the interactive features:
 \begin{equation}
		 \hat{\mathcal{Q}}_{b} = \mathcal{I}(Q= \mathcal{Q}_{b},K= V= \hat{E}) 
 \end{equation}
 Subsequently, for each interactive behavior query $\hat{\mathcal{Q}}_{b}$, 
 interactions between different modalities are parsed utilizing a self-attention module.
 \begin{equation}
	 \hat{\mathcal{Q}}_{b,mode} = \mathcal{I}(Q= K= V=\hat{\mathcal{Q}_{b}}) 
 \end{equation}
 Unlike previous algorithms that only utilize goals to guide predictions, 
 an MLP decodes the behavior queries to initial trajectory 
 proposals $\hat{\mathcal{Y}}_{coarse} \in \mathbb{R}^{T_f \times K \times 2}$. 
 We follow the approach of previous algorithms and use a Winner-Takes-All training 
strategy,
 an auxiliary $L_2$ loss $\mathcal{L}_{behav}$ is used to supervise 
 the quality of the generated initial trajectory proposal: 
 \begin{equation}
	\mathcal{L}_{behav} = \min_{k\in \left \{ 1,2,...,K \right \} }\frac{1}{T_f} \sum_{t=1}^{T_f}  \left \|\hat{\mathcal{Y}}^{t,k}_{coarse} - \mathcal{Y}^t \right \|_2^2
\end{equation}
where $\hat{\mathcal{Y}}^k_{coarse}$ denotes $\hat{\mathcal{Y}}_{coarse}$ for the $k-th$ modality.
Finally, each proposal point is concatenated with the behavior
query of its corresponding modality to obtain the final
behavior representations
$\hat{\mathcal{Q}}_{b,out}\in\mathbb{R}^{K\times T_f\times C_b}$,
where the original mode-level behavior queries are expanded
into mode-time representations that simultaneously encode the
corresponding motion mode and prediction timestamp. In this
way, multimodal future motion hypotheses can be modeled using
only $K$ learnable behavior tokens, avoiding the need to
explicitly construct $K\times T_f$ behavior queries.

\subsection{Lane Attention Branch}
For vehicles traveling in structured road environments,
future trajectories are not only influenced by driving intentions
but are also constrained by lane topology.
Moreover, lane constraints may vary throughout the prediction horizon,
as different lane segments can influence vehicle motion at different
future moments. Therefore, providing a single lane-level representation
may be insufficient to characterize the evolving lane constraints
associated with future trajectories.

To address this issue, we introduce a lane attention branch
that provides fine-grained lane topology priors throughout the
prediction horizon. Specifically, timestamp-level lane tokens are
designed to capture lane-related constraints that may influence
future motion at different prediction timestamps. Through interaction
with lane features and supervision from future lane occupancy
information, the proposed lane tokens learn latent lane topology
representations relevant to future trajectory generation.

First, $T_f$ learnable tokens $L_{tokens}$ are assigned to the
target encoding $\hat{E}_{tar}$ to obtain the initial lane queries
$\mathcal{Q}_L \in \mathbb{R}^{T_f \times C_e}$:

\begin{equation}
\mathcal{Q}_L = \hat{E}_{tar}+L_{tokens}
\end{equation}
 Then, these lane queries are fed into a cross attention module to aggregate comprehensive 
 scene features:
 \begin{equation}
 \mathcal{\hat{Q}}_L  = \mathcal{I} \left ( Q=\mathcal{Q}_L,K=V=\hat{E}  \right ) 
 \end{equation}

 In the previous sections, we have obtained the encoded information $\hat{E}_\mathcal{M}$  corresponding to each lane segment.
 We want each $\mathcal{\hat{Q}}_L$ to aggregate the lane segment features
  that will be reached at its corresponding future moment.
 Inspired by \cite{liu2024laformer,mo2023map}, a lane scoring module is designed to 
predict the lane probabilities at each future moment. 
Specifically, we first treat each  $\mathcal{\hat{Q}}_L$ as the key and value, 
map encoding as queries, and compute the query-lane interaction features $\mathcal{I} _{\mathcal{L} ,\mathcal{M} }$. 
The $\mathcal{I} _{\mathcal{L} ,\mathcal{M} }$, map encoding, and interactive lane queries  are concatenated and fed into an attention mechanism to compute the query-lane scores:

\begin{equation}
\mathcal{S}^{m,t_f} _{\mathcal{L} ,\mathcal{M} } = \frac{\exp(\mathrm{MLP} (\mathcal{I}^{m,t_f} _{\mathcal{L},\mathcal{M} },\hat{E}^m_{\mathcal{M}},\mathcal{\hat{Q}}^{t_f}_L))}{\sum_{n=1}^{N_m} \exp(\mathrm{MLP} (\mathcal{I}^{n,t_f} _{\mathcal{L},\mathcal{M} },\hat{E}^n_{\mathcal{M}},\mathcal{\hat{Q}}^{t_f}_L))} 
\label{scores}
\end{equation}
where $\mathcal{S}^{m,t_f} _{\mathcal{L} ,\mathcal{M} }$ denotes the attention
 score of $t_f$-th lane query for the $m$-th lane segment. 
We utilize a binary cross-entropy lane 
loss to supervise the attention scores:
\begin{equation}
\mathcal{L}_{lane } = \sum_{t_f=1}^{T_f}  \mathcal{L} _{CE} \left ( \mathcal{S}^{t_f} _{\mathcal{L} ,\mathcal{M} },\mathcal{S}^{t_f} _{GT} \right ) 
\end{equation}
where we assign a ground truth label of 1 to the nearest lane segment that the target vehicle arrives at the moment $t_f$, and a label of 0 to the rest of the lane segments.  
We select and concatenate the top $M$ lane segments features with the $M$ highest
 attention scores for each query to obtain $L^{t_f}_{con}$:
\begin{equation}
L^{t_f}_{con} = \mathrm{Concat}\left ( top_M(\hat{E} _{\mathcal{M}}),top_M({\mathcal{S}^{t_f}}) \right ) 
\end{equation}
where $top_M()$ denotes the selection function.
$L^{t_f}_{con}$ are treated as the keys and values to be aggregated into each lane query:
\begin{equation}
	\begin{split}
\hat{\mathcal{Q} }^{t_f}_{lane} = \mathcal{I}(Q = \mathcal{\hat{Q}}^{t_f}_{L},K = V = L^{t_f}_{con}) 
	\end{split}
\end{equation}
Under the supervision of the lane loss, our interactive lane
queries are aligned with the lane segments that will be reached
at the corresponding future moments, thus effectively guiding
the prediction. Since each lane query aggregates the top-$M$
most probable lane segments, it already captures multimodal
lane topology information associated with the corresponding
future timestamp. To facilitate fusion with the behavior-aware
representations, the lane features are expanded using a simple MLP along the modal
dimension to obtain lane-aware representations
$\hat{\mathcal{Q}}_{lane}\in\mathbb{R}^{K\times T_f\times C_l}$,
which share the same mode-time structure as
$\hat{\mathcal{Q}}_{b,out}$.

\subsection{Two-stage Decoder}
\begin{figure}
	\centering
	\includegraphics[width=8cm]{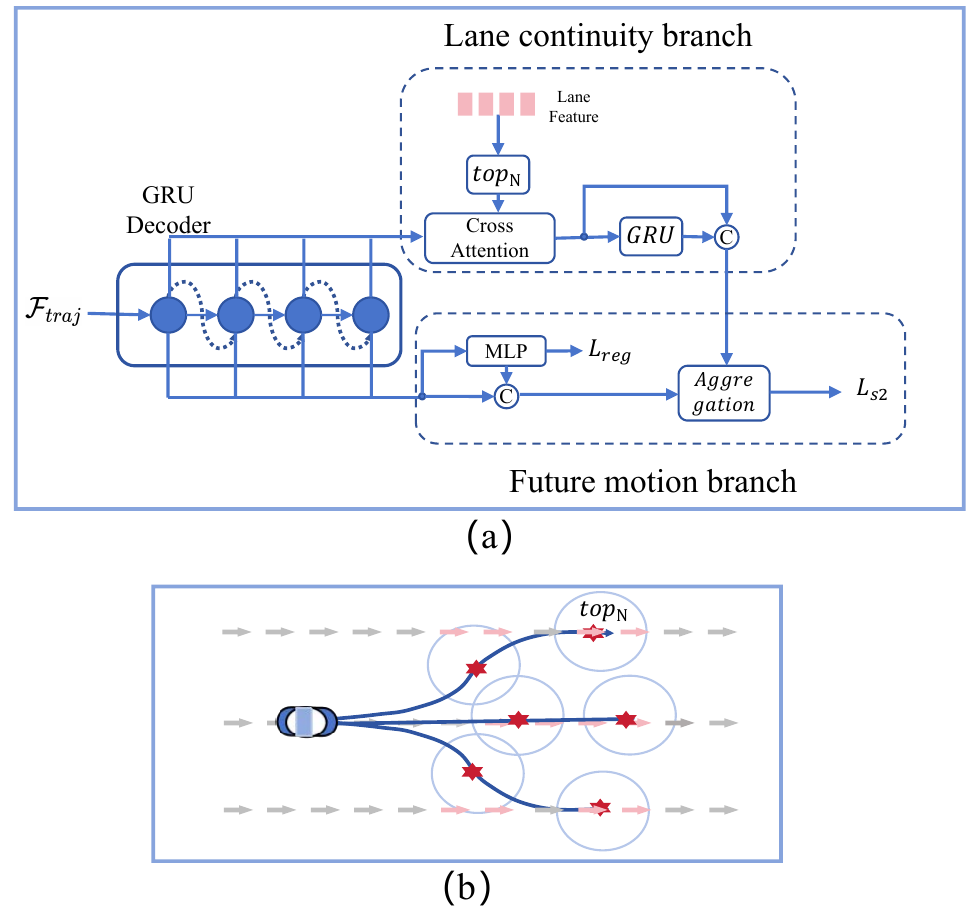}
	\caption{(a) The two-stage decoder. We first use a simple GRU decoder to 
	predict the target proposals. Subsequently, we use the lane continuity 
	and future behavior to refine the proposals.
	 (b) Illustration of distance-based lane segments selection.} 
	\label{fig:decoders}
\end{figure}

This section presents a two-stage decoder as shown in Fig. \ref{fig:decoders} (a).
In the first stage, it predicts the trajectories of the target vehicles and the modal probabilities. 
Subsequently, In the second stage, the 
continuity of the lane constraints and future motion features are
 utilized to refine the trajectories to obtain the final prediction, 
 the modal probabilities are also updated.
 For better presentation, a detailed pseudo-code for the two-stage decoder 
 is given in Algorithm \ref{alg:work_flow}.

 \begin{algorithm}[htpb]
	\caption{The two-stage decoder}
	\label{alg:work_flow}
	\KwIn{ \\
	\ \ \ \ \ $\hat{\mathcal{Q}}_{b,out}$, 
	$\hat{\mathcal{Q} }_{lane}$, 
	$\hat{E}_\mathcal{M}$ and lane segment positions $P_\mathcal{M}$; 
}
	\KwOut{ \\
		\ \ \ \ \ Predicted trajectories $\hat{\mathcal{Y}}$ and probabilities $\hat{P}$;
	}
	\BlankLine
	
	\SetKwFunction{FMain}{Lanecon}

	Aggregate $\hat{\mathcal{Q}}_{b,out}$ and $\hat{\mathcal{Q}}_{lane}$ 
	at $T_f$ timestamp to predict modal probabilities $\hat{P}$.\\

	Aggregate $\hat{\mathcal{Q}}_{b,out}$ and $\hat{\mathcal{Q} }_{lane}$ and encode 
	them to obtain future prediction features $\mathcal{F}_{fut}$  using GRU according to \eqref{mlpagg} and \eqref{GRU1};
  
	Decode $\mathcal{F} _{fut}$ to obtain predicted multimodal proposals $\hat{\mathcal{Y}}$, 
	and $\hat{\mathcal{Y}_{\mu}}$ is the position of the predicted proposals;

	\SetKwFunction{FMain}{Lanecon}
	
	\If{perform proposals refinement}{
		Obtain future motion features $\mathcal{F}_{b}$ via \eqref{behav} \\
		$\mathcal{F}'_{Lanecon} \leftarrow$ \FMain($\mathcal{F}_{fut}, \hat{\mathcal{Y}}_{\mu}, \hat{E}_\mathcal{M}, P_\mathcal{M}$) \\
		Predict $\Delta\hat{\mathcal{Y}}_{\mu}$ by aggregating $\mathcal{F}_{b}$ and $\mathcal{F}'_{Lanecon}$ \\
		$\hat{\mathcal{Y}}_{\mu} \leftarrow \hat{\mathcal{Y}}_{\mu} + \Delta\hat{\mathcal{Y}}_{\mu}$ \\
		Update $\hat{P}_{new}$ with $\mathcal{F}_{b}$ and $\mathcal{F}'_{Lanecon}$ at $T_f$ \\
		$\hat{P} \leftarrow \hat{P}_{new}$ \\
		\Return $\hat{\mathcal{Y}}$ and $\hat{P}$
	}
	\Else{
		\Return $\hat{\mathcal{Y}}$ and $\hat{P}$ \\
	}
	\BlankLine
	\SetKwProg{Fn}{function}{:}{}
	\Fn{\FMain{$\mathcal{F}_{fut}, \hat{\mathcal{Y}_{\mu}}, \hat{E}_\mathcal{M}, P_\mathcal{M}$}}{
    \For{$k \leftarrow 1$ \KwTo $K$}{
        \For{$t \leftarrow 1$ \KwTo $T_f$}{
            $\texttt{dist} \leftarrow \|\hat{\mathcal{Y}_{\mu}}^{t,k} - P_\mathcal{M}\|_2^2$ \\
            $\texttt{idxs} \leftarrow \arg\min^N(\texttt{dist})$ \\
            $\hat{E}_\texttt{sel} \leftarrow \hat{E}_\mathcal{M}[\texttt{idxs}]$ \\
            Aggregate $\hat{E}_\texttt{sel}$ to obtain $\mathcal{F}_{lane}^{t,k}$ according to \eqref{lane}  \\
        }
    }
	Using GRU to preserve lane continuity to get $\mathcal{F}'_{Lanecon}$ according to \eqref{cont}.\\
    \Return $\mathcal{F}'_{Lanecon}$ \\
}

	\textbf{end function}
\end{algorithm}

\subsubsection{GRU-based predictor}

In the first stage, we predict future trajectories using a
Laplacian mixture density decoder. Specifically, the
behavior-aware representations $\hat{\mathcal{Q}}_{b,out}$ and
lane-aware representations $\hat{\mathcal{Q}}_{lane}$ are
concatenated and projected by an MLP to obtain the trajectory
features $\mathcal{F}_{traj}\in\mathbb{R}^{K\times T_f\times C_{traj}}$:
\begin{equation}
\mathcal{F}_{traj}
=
\mathrm{MLP}
\left(
\mathrm{Concat}\left(
\hat{\mathcal{Q}}_{lane},
\hat{\mathcal{Q}}_{b,out}
\right)
\right).
\label{mlpagg}
\end{equation}
In this way, future trajectory generation is jointly guided by
latent behavioral evolution patterns and temporally varying lane
topology constraints.

The target encoding $\hat{E}_{tar}$ is then used as the initial
hidden state of a GRU to recover future motion features:
\begin{equation}
\mathcal{F}_{fut}
=
\mathrm{GRU}
\left(
\mathcal{F}_{traj},
h_0=\hat{E}_{tar}
\right).
\label{GRU1}
\end{equation}

The recovered features $\mathcal{F}_{fut}$ are fed into two
stacked MLPs to predict the location and scale parameters of
future trajectories. Meanwhile, the final-step behavior-aware
and lane-aware representations of each modality are used to
estimate the corresponding mode probabilities.

The multimodal prediction results are formulated as:
\begin{equation}
    \hat{\mathcal{Y}} = \sum_{k=1}^{K} \hat{p}_k \mathrm{Laplace} \left( \hat{\mathcal{Y}}_{\mu}^k, \hat{\mathcal{Y}}_{\sigma}^k \right),
\end{equation}
where $\hat{p}_k$ denotes the probability of the $k$-th mode
and satisfies $\sum_{k=1}^{K}\hat{p}_k=1$.
$\hat{\mathcal{Y}}_{\mu}\in\mathbb{R}^{T_f\times K\times 2}$
and
$\hat{\mathcal{Y}}_{\sigma}\in\mathbb{R}^{T_f\times K\times 2}$
denote the predicted location and scale parameters of the
corresponding Laplace distributions, respectively.

\subsubsection{Trajectory refinement}
In the second stage, we try to refine the predictions at the point-level.
While fine-grained future behavioral intentions and lane constraints are
provided in dual-stream attention networks, the uncertainty of the forecast still exists.
This is mainly due to the fact that the prediction results depend
 heavily on the quality of the predicted lane and behavioral constraints.
When the predicted lane constraints from the lane attention branch run counter to the multi-modal 
behavioral intentions provided by the behavior state attention branch at a certain time,
 some unreasonable proposals can be produced.
 To this end, we design a point-level refinement method
using the continuity of lane constraints and future motion features.

Specifically, we first represent future motion features $\mathcal{F}_{b}$ by concatenating the 
corresponding future prediction features with the proposal positions.
\begin{equation}
	\mathcal{F}_{b}^{t_f} = \mathrm{Concat}( \mathcal{F}_{fut}^{t_f},\hat{\mathcal{Y}}^{t_f} ) 
	\label{behav}
\end{equation}
Inspired by the fact that the lane segments that vehicles travel through are 
continuous, we try to model the lane continuity into the network.
 For each trajectory point, we filter and aggregate the nearest $N$ lane segment features as shown in Fig. \ref{fig:decoders} (b):
 \begin{equation}
	\mathcal{F}^{t,k}_{lane}  = \mathcal{I} \left ( Q=\mathcal{F}^{k,t_f} _{fut},
	K=V=top_N(\hat{E}_\mathcal{M})  \right ) 
\label{lane}
\end{equation}
where $\mathcal{F}^{t,k}_{lane}$ denotes the aggregated lane feature for trajectory points 
at moment $t$ of $k-th$ mode.  Subsequently, we utilize a GRU to encode $\mathcal{F}_{lane}$ in the temporal dimension 
to preserve the continuity of the passing lanes:
 \begin{equation}
	\mathcal{F}'_{Lanecon}  = GRU( \mathcal{F}_{lane})
	\label{cont}
 \end{equation}
 Finally, future behavioral features $ \mathcal{F}_{b}$ and lane continuity features $\mathcal{F}'_{Lanecon}$ are aggregated to 
 predict the proposal deviations $\Delta \mathcal{Y} = \mathcal{Y} - \hat{\mathcal{Y} }$ between proposals and ground truth. 
 Modal probabilities are also re-predicted, using the last-moment future behavioral features combined with lane continuity features.

\subsection{Model Training}
We train the network in two stages. The first stage of the 
decoder is first trained by minimizing the regression loss $\mathcal{L}_{reg}$ 
and classification loss $\mathcal{L}_{cls}$ of the two auxiliary losses.
\begin{equation}
\mathcal{L}_{s1}  =\mathcal{L}_{reg}+\mathcal{L}_{cls}+
\lambda_L\mathcal{L}_{lane}+\mathcal{L}_{behav}
\end{equation}
where $\lambda_L$ is the lane loss weight. 
The negative log-likelihood estimation 
is used to compute regression loss $\mathcal{L}_{reg}$. 
We follow the approach of previous algorithms and use a Winner-Takes-All training 
strategy to focus on the mode $k^*$ with the minimum average Euclidean 
distance to the ground truth among the K proposals:
\begin{equation}
\mathcal{L}_{reg} = -\frac{1}{T_f} \sum_{t=1}^{T_f} \log P \left( \mathcal{Y}^{t} \;\middle|\; \hat{\mathcal{Y}}_\mu ^{t,k^*}, \hat{\mathcal{Y}}_\sigma^{t,k^*} \right)
\end{equation}
    In addition, inspired by \cite{zhou2022hivt}, a soft displacement error-based cross-entropy loss is adopted for multimodal trajectory classification:
    \begin{equation}
        \mathcal{L}_{cls} = -\sum_{k=1}^{K} p_k \log(\hat{p}_k),
    \end{equation}
    where $p_k$ is the soft target probability for the $k$-th trajectory mode, obtained by applying a softmax operation to the negative average displacement error (ADE):
    \begin{equation}
        p_k = \frac{\exp(-d_k)}{\sum_{j=1}^{K}\exp(-d_j)},
    \end{equation}
    where $d_k$ denotes the ADE between the $k$-th predicted trajectory mode and the ground truth:
    \begin{equation}
        d_k = \frac{1}{T_f} \sum_{t=1}^{T_f} \left\| \hat{\mathcal{Y}}^{t,k}_{\mu} - \mathcal{Y}^t \right\|_2.
    \end{equation}

In the second training stage, all network components,
including the encoder, dual-stream attention branches,
first-stage decoder, and refinement module, remain trainable
and are jointly optimized in an end-to-end manner.
The loss function extends that of the first stage by additionally
incorporating a deviation loss $\mathcal{L}_{d}$ and a heading angle
loss $\mathcal{L}_{angle}$:
\begin{equation}
	\mathcal{L}_{s2}
	=
	\mathcal{L}_{s1}
	+
	\lambda_d \mathcal{L}_{d}
	+
	\lambda_a \mathcal{L}_{angle}.
\end{equation}
Similar to the first stage of training, we utilize Winner-Takes-All's 
strategy to compute the deviation 
loss and heading angle loss for $k*$-modal proposals:
\begin{equation}
	\mathcal{L}_{d} = \frac{1}{T_f} \sum_{t=1}^{T_f}  \left\|\Delta\mathcal{ \hat{Y}}^{ k^*,t} - \Delta\mathcal{Y}^{ k^*,t} \right \|_2^2
\end{equation}
\begin{equation}
	\mathcal{L}_{angle} = -\frac{1}{T_f} \sum_{t=1}^{T_f} \cos\left (\hat{\theta} ^{ k^*,t} -\theta^t \right ) 
\end{equation}
where $\Delta\mathcal{ \hat{Y}}^{ k^*,t}$ denotes the predicted deviation for the k*-mode.
$\hat{\theta} ^{ k^*,t}$ and $\theta ^{t}$  represent the heading angle of 
the trajectory of the $k*$-mode and the ground truth heading angle at moment $t$, respectively.

\section{Experiments}
\subsection{Experiments Setup}

\subsubsection{Datasets} We validate the performance of our algorithm on
 two publicly available large datasets, nuScenes \cite{caesar2020nuscenes} and Argoverse \cite{chang2019argoverse}.

\textbf{Argoverse} motion forecasting dataset contains a variety of compelling scenarios
 extracted from 1,006 driving hours in Miami and Pittsburgh, totaling 333,441 five-second
  sequences. The prediction task involves forecasting the subsequent three-second trajectory
   of the target agent based on its own trajectory and those of neighboring agents in the 
   initial two seconds. 

\textbf{nuScenes} is a dataset that features a wide array of complex lane scenes captured in urban environments in Boston and Singapore. It includes over 1000 driving scenarios, each lasting 20 seconds. The dataset is primarily focused on the prediction task, where the goal is to forecast  trajectories for the upcoming 6 seconds based on observations from the preceding 2 seconds.

\subsubsection{Evaluation Metrics}

To evaluate the fit of the multimodal trajectories to the ground truth (GT), we utilize commonly used evaluation metrics: 

\textbf{Minimum average displacement error ($minADE_K$)}:  the minimum of the mean L2 distance of the predicted k multimodal trajectories from GT:

\begin{equation}
	minADE_K = \min \left( ( \frac{1}{T_f} \sum_{t=1}^{T_f}  
	\left \| \hat{\mathcal{Y}}_\mu ^{t,k} -\mathcal{Y}^t   \right \| _2 \right ) \quad  k=1,2,...,K 
\end{equation}

\textbf{Minimum final displacement error} ($minFDE_K$): the minimum value of the L2 distance between the endpoints of the predicted k multimodal trajectories and the GT endpoint:

\begin{equation}
	minFDE_K =  \min \left(  \left \| \hat{\mathcal{Y}}_\mu ^{T_f,k} -\mathcal{Y}^{T_f}   \right \| _2 \right ) \quad  k=1,2,...,K 
\end{equation}
When $K = 1$ in the above evaluation metrics, it indicates the calculation of ADE
 and FDE for the most likely trajectories.

\textbf{ Brier minimum Final Displacement Error}  ($b-minFDE_K$): This metric is similar to $minFDE_K$, but
additional $(1.0 -\hat{p}_K)^2$  is added to the endpoint loss.

\subsection{Quantitative Analysis}
In this section, we quantitatively compare the performance of the proposed model with 
other state-of-the-art algorithms. 

\subsubsection{Comparison with State-of-the-Arts in NuScenes} 

\begin{table*}[h]
	\centering
	\caption{Comparison with other state of the art algorithms in the
	 NuScenes dataset leaderboard. $\downarrow$: smaller is better.}
	\label{table:nu}
\begin{tabular}{l l c c c c c c}
\toprule
Model & Year & minFDE$_1$$\downarrow$ & minADE$_5$$\downarrow$ & minFDE$_5$$\downarrow$ & minADE$_{10}$$\downarrow$ & minFDE$_{10}$$\downarrow$ & Param. (M)$\downarrow$ \\
\midrule
CoverNet \cite{phan2020covernet} & 2020 & 9.26 & 1.96 & -- & 1.48 & -- & 5.7 \\
Trajectron++\cite{salzmann2020trajectron++} & 2020 & 9.52 & 1.88 & -- & 1.51 & -- & 0.3 \\
Lapred \cite{kim2021lapred} & 2021 & 8.12 & 1.53 & 3.37 & 1.12 & 2.39 & 1.8 \\
PGP\cite{deo2022multimodal} & 2021 & 7.17 & 1.27 & 2.47 & 0.95 & 1.55 & 0.1 \\
AutoBot\cite{girgis2021latent} & 2022 & 8.19 & 1.37 & -- & 1.03 & -- & 1.5 \\
GOHOME\cite{gilles2022gohome} & 2022 & 6.99 & 1.42 & -- & 1.15 & -- & 0.4 \\
THOMAS \cite{gilles2021thomas} & 2022 & 6.71 & 1.33 & -- & 1.04 & -- & -- \\
ContextVAE\cite{xu2023context} & 2023 & 8.24 & 1.59 & 3.28 & -- & -- & 11.7 \\
Caspformer\cite{yadav2024caspformer} & 2024 & 6.70 & 1.15 & -- & -- & -- & -- \\
Caspnet++\cite{schafer2024caspnet++} & 2024 & 6.18 & 1.18 & -- & 0.93 & -- & -- \\
Efficient\cite{li2024efficient} & 2024 & 7.43 & 1.34 & -- & -- & -- & 0.4 \\
Goal-LBP\cite{yao2023goal} & 2024 & -- & 1.02 & 1.87 & 0.93 & 1.65 & -- \\
G2LTraj\cite{zhang2024g2ltraj} & 2024 & 8.30 & 1.40 & -- & 0.96 & -- & 0.3 \\
SemanticFormer\cite{sun2024semanticformer} & 2024 & 6.27 & 1.14 & -- & 0.87 & -- & -- \\
Unitraj\cite{feng2024unitraj} & 2024 & -- & 0.96 & -- & 0.84 & -- & 60.1 \\
FIM\cite{pei2025foresight} & 2025 & -- & 0.88 & -- & 0.79 & -- & -- \\
GC-GAT\cite{gulzar2025gc} & 2025 & -- & 1.19 & -- & 1.06 & -- & -- \\
Diffutory\cite{lan2025diffutory} & 2025 & -- & 1.21 & 2.32 & 0.97 & 1.64 & 1.80 \\
\textbf{Our Method} & -- & 6.67 & 1.16 & 2.18 & 0.91 & 1.44 & 0.6 \\
\bottomrule
\end{tabular}
\end{table*}
As summarized in Table \ref{table:nu}, we comprehensively compare our proposed method with a collection of state-of-the-art trajectory prediction approaches published from 2020 to 2025 on the official NuScenes benchmark, covering classic baselines and recent top-performing solutions on the public leaderboard. Our method  attains comparable prediction accuracy with mainstream approaches proposed between 2022 and 2024, including GOHOME (2022), THOMAS (2022), CaspNet++ (2024)  and CaspFormer (2024). In terms of the top-10 multimodal prediction 
metrics, our method outperforms GC-GAT published in IEEE RAL 2025 and
 Diffutory  published in IEEE TITS 2025. Specifically, our method
  achieves $\text{minADE}_{10}$ of 0.91 and $\text{minFDE}_{10}$ of 1.44, 
  which surpass GC-GAT ($\text{minADE}_{10}=1.06$) and 
  Diffutory ($\text{minADE}_{10}=0.97$), and also outperforms
   Goal-LBP  on these two metrics. Although our performance is
     inferior to SemanticFormer,
	 our approach still exhibits strong overall competitiveness 
	 and achieves acceptable prediction accuracy.
Unitraj (2024)  obtains the leading numerical results across several evaluation metrics. However, Unitraj is trained on a large-scale unified dataset containing 2 million trajectories aggregated from nuScenes, Argoverse and Waymo Open Motion Dataset, while our model and all other baselines are trained solely on the NuScenes dataset without cross-dataset data expansion. Notably, our method achieves far better prediction results than FIM on the Argoverse dataset, which demonstrates the stronger generalization performance of our framework across different traffic scenarios.

\subsubsection{Comparison with State-of-the-Arts in Argoverse}

\begin{table}[h]
	\centering
	\caption{Comparison with other state of the art algorithms in the Argoverse dataset leaderboard. $\downarrow$: smaller is better. }
	\label{table:arg}
	\resizebox{\columnwidth}{!}{%
\begin{tabular}{l l c c c c c}
\toprule
Model & Year & minADE$_6$$\downarrow$ & minFDE$_6$$\downarrow$ & b-minFDE$_6$$\downarrow$ & Param. (M)$\downarrow$ \\
\midrule
LaneGCN \cite{liang2020learning} & 2020 & 0.87 & 1.36 & 2.05 & 3.7 \\
SceneTrans \cite{ngiam2021scene} & 2022 & 0.80 & 1.23 & 1.90 & -- \\
HiVT \cite{zhou2022hivt} & 2022 & 0.77 & 1.17 & 1.84 & 2.3 \\
LTP \cite{wang2022ltp} & 2022 & 0.83 & 1.30 & 1.86 & 1.1 \\
Adapt \cite{aydemir2023adapt} & 2023 & 0.79 & 1.17 & 1.80 & 1.4 \\
GANet \cite{wang2023ganet} & 2023 & 0.80 & 1.16 & 1.79 & 5.0 \\
PBP \cite{afshar2024pbp} & 2023 & 0.86 & 1.33 & 1.98 & -- \\
QCNet \cite{zhou2023query} & 2023 & 0.73 & 1.07 & 1.69 & 7.7 \\
Prophnet \cite{wang2023prophnet} & 2023 & 0.77 & 1.14 & 1.73 & 9.3 \\
FFiNet \cite{kang2024ffinet} & 2024 & 0.76 & 1.12 & 1.73 & 6.2 \\
HPNet \cite{tang2024hpnet} & 2024 & 0.76 & 1.10 & 1.74 & 4.1 \\
HHLF \cite{jiao2024hierarchical} & 2024 & 0.81 & 1.25 & -- & -- \\
Multipath++ \cite{varadarajan2022multipath++} & 2022 & 0.79 & 1.21 & 1.79 & -- \\
Q-EANet \cite{chen2024q} & 2024 & 0.80 & 1.23 & 1.92 & 0.8 \\
SEPT \cite{lan2023sept} & 2024 & 0.73 & 1.06 & 1.69 & -- \\
SimpL \cite{zhang2024simpl} & 2024 & 0.79 & 1.18 & 1.81 & 1.6 \\
FutureNet-LOF \cite{wang2025futurenet} & 2025 & 0.73 & 1.03 & 1.66 & -- \\
VCIFormer \cite{li2025toward} & 2025 & 0.75 & 1.28 & -- & 1.3 \\
FIM \cite{pei2025foresight} & 2025 & 0.83 &  1.21 & 1.83 & -- \\
GoIRL \cite{pei2025goirl} & 2025 & 0.81 &  1.17 & 1.80 & -- \\
\textbf{Our Method} & -- & 0.75 & 1.11 & 1.78 & 2.3 \\
\bottomrule
\end{tabular}
}
\end{table}

Similar to the experiments conducted on the NuScenes dataset, we further 
validate the generalization performance of our proposed method on 
the Argoverse dataset, and the quantitative comparison results 
against various state-of-the-art approaches are summarized in
 Table \ref{table:arg}. We select representative trajectory 
 prediction algorithms published from 2020 to 2025 for comparison,
  covering classic baseline models and recent advanced methods
   proposed in top conferences and journals.
Our method achieves competitive performance on three mainstream evaluation metrics including $\text{minADE}_6$, $\text{minFDE}_6$ and $\text{b-minFDE}_6$. Our approach outperforms several recently proposed methods such as HHLF and VCIFormer. Specifically, we obtain a much lower $\text{minFDE}_6$ than VCIFormer.
We further compare our method with FIM and GoIRL published in 2025. Our method achieves $\text{minADE}_6=0.75$ and $\text{minFDE}_6=1.11$, which outperforms FIM ($\text{minADE}_6=0.83$, $\text{minFDE}_6=1.21$) and GoIRL ($\text{minADE}_6=0.81$, $\text{minFDE}_6=1.17$) as reported in their original papers.
A few approaches including QCNet, SEPT and FutureNet-LOF achieve marginally 
better prediction accuracy via elaborate network designs. However, 
our method still demonstrates strong 
competitiveness.

\subsection{Ablation Study}
In order to analyze the contribution of each network module and determine the optimal 
network structure, we conduct extensive ablation experiments.
\subsubsection{Contribution of each network component}
The prediction performance of different network variants is summarized in Table \ref{table:ab1}, which systematically 
demonstrates the contribution of each component to the overall framework.

\textbf{1. Baseline with Encoder Features (Row 1): } Using encoder features alone 
generates rough predictions due to the lack of the guidance of the priors.

\textbf{2. Behavior State Attention Branch (Row 2):} Incorporating the behavior state attention branch introduces fine-grained driving intentions, therefore the prediction accuracy is boosted.

\textbf{3. Lane Attention Branch (Row 4):} The introduction of the lane attention branch further brings fine-grained lane constraints at each moment to the network to jointly guide the prediction and improve the prediction performance.

\textbf{4. Decoder Design Comparison (Row 3):} Replacing the GRU decoder with an LSTM leads to suboptimal results, indicating that GRU's simpler architecture is more effective for trajectory decoding in this framework.

\textbf{5. Refinement Module (Rows 5-6):}

\begin{itemize}
	\item Adding future motion refinement (Row 5) reduces the $minFDE$ metric (especially for $minFDE_1$)
	by aligning predictions with intent-aware features.
	
	\item Further integration of the lane continuity constraint (Row 6) ensures the continuity of the trajectory, which in turn leads to the optimization of all evaluation metrics.
\end{itemize}

\begin{table*}[h]
	\centering
	\caption{Ablation Study of Network Components}
	\label{table:ab1}
	\begin{tabular}{cccccc|ccc}
	\toprule[1pt]
	\multirow{2}{*}{\textbf{Encoder}} & 
	\multirow{2}{*}{\begin{tabular}[c]{@{}c@{}}\textbf{Behavior State Attention}\end{tabular}
	} & 
	\multirow{2}{*}{
	\begin{tabular}[c]{@{}c@{}}\textbf{Lane Attention}\end{tabular}
	} & 
	\multirow{2}{*}{
	\begin{tabular}[c]{@{}c@{}}\textbf{Decoder Type}\end{tabular}
	} & 
	\multicolumn{2}{c|}{\textbf{Refinement}} & 
	\multirow{2}{*}{\textbf{minFDE\textsubscript{1}}} & 
	\multirow{2}{*}{\textbf{minADE\textsubscript{5}}} & 
	\multirow{2}{*}{\textbf{minFDE\textsubscript{5}}} \\
	\cline{5-6}
	 & & & & 
	\textbf{
	\begin{tabular}[c]{@{}c@{}}Future\\Motion\end{tabular}
	} & 
	\textbf{
	\begin{tabular}[c]{@{}c@{}}Lane\\Continuity\end{tabular}
	} & & & \\
	\midrule[0.8pt]
	
	$\checkmark$ & -- & -- & GRU & -- & -- & 7.52 & 1.26 & 2.50 \\
	$\checkmark$ & $\checkmark$ & -- & GRU & -- & -- & 7.28 & 1.21 & 2.36 \\
	$\checkmark$ & $\checkmark$ & $\checkmark$ & LSTM & -- & -- & 7.12 & 1.19 & 2.31 \\
	$\checkmark$ & $\checkmark$ & $\checkmark$ & GRU & -- & -- & 6.99 & 1.17 & 2.26 \\
	$\checkmark$ & $\checkmark$ & $\checkmark$ & GRU & $\checkmark$ & -- & 6.84 & 1.17 & 2.23 \\
	$\checkmark$ & $\checkmark$ & $\checkmark$ & GRU & $\checkmark$ & $\checkmark$ & \textbf{6.67} & \textbf{1.16} & \textbf{2.18} \\
	\bottomrule[1pt]
	\end{tabular}
	\end{table*}

\subsubsection{Impact of different numbers of lane segments}
In the Lane Attention Branch, top $M$ lane segments with the highest probability scores are selected.
We explore the effect of different number of lane segments $M$.
Table \ref{table:ab2} shows the performance of our network with different number of lane segments $ M=\left \{ {1, 2, 3,4} \right \}$ .
It can be seen that $M=2$ achieves peak performance in multimodal-trajectory prediction 
(minADE\textsubscript{5}: 1.17, minFDE\textsubscript{5}: 2.26), while $M=1$ attains the lowest minFDE\textsubscript{1}.
 Notably,  further increasing $M$ to 4 yields stagnant performance  with marginally higher minFDE\textsubscript{1}. 
 This suggests that moderate lane selection ($M=2$) optimally balances feature sufficiency, 
 whereas excessive segments ($M \geq 3$) introduce redundant features without predictive benefits.

 Similarly, in the Trajectory Refinement, nearest $N$ lane segment features are selected
 for each trajectory point. From the prediction results in Table \ref{table:abrefine},
  we can conclude that  selecting the nearest $N=2$ lane segments yields optimal prediction 
  performance across all metrics. While increasing $N$ from 1 to 2 improves trajectory
   precision, further expansion to $N \geq 3$ introduces redundant lane features that
    dilute the model's focus, resulting in consistent performance degradation.

 \begin{table}[]
	\centering
	\caption{Ablation study on the number of most probable lane segments $M$}
	\label{table:ab2}
			\resizebox{\columnwidth}{!}{%
			\begin{tabular}{clcc}
				\toprule
				\begin{tabular}[c]{@{}c@{}} \textbf{The number of the} \\ \textbf{Selected Lane Segments M}\end{tabular} & $\textbf{minFDE}_1$    & $\textbf{minADE}_5$    & $\textbf{minFDE}_5$    \\ \midrule
				1                                                                                     & \textbf{6.90} & 1.18          & 2.27          \\
				2                                                                                     & 6.99          & \textbf{1.17} & \textbf{2.26} \\
				3                                                                                     & 7.01          & 1.18          & 2.27          \\
				4                                                                                     & 7.00          & 1.18          & 2.27          \\ \bottomrule
				\end{tabular}
}
\end{table}

\begin{table}[]
	\centering
	\caption{Ablation study on the number of nearest line segment $N$}
	\label{table:abrefine}
			\resizebox{\columnwidth}{!}{%
			\begin{tabular}{clcc}
				\toprule
				\begin{tabular}[c]{@{}c@{}} \textbf{The number of the} \\ \textbf{Selected Lane Segments N}\end{tabular} & $\textbf{minFDE}_1$    & $\textbf{minADE}_5$    & $\textbf{minFDE}_5$    \\ \midrule
				1                                                                                     & 6.83 & 1.16          & 2.18          \\
				2                                                                                     &\textbf{6.67}          & \textbf{1.16} & \textbf{2.18} \\
				3                                                                                     & 6.92          & 1.17          & 2.21          \\
				4                                                                                     & 6.93          & 1.17          & 2.21          \\ \bottomrule
				\end{tabular}
}
\end{table}
\subsubsection{Effect of fine-grained queries}
In contrast to previous algorithms that used only the goal prior to guide the entire trajectory prediction, 
our algorithm employs fine-grained queries.
Therefore, we further explore the advantages of our proposed method, 
as shown in the Table \ref{table:ab3}.
For a fair comparison, predictions guided by both fine-grained queries and
 only goal queries are not refined. 
 It can be seen that the fine-grained queries achieve
  improvements over goal-based queries, with lower prediction errors across all 
  key metrics ($minFDE_1$, $minADE_5$, $minFDE_5$). This highlights the effectiveness of 
  fine-grained queries in capturing nuanced spatiotemporal details, enabling more precise trajectory predictions.

\begin{table}
		\centering
	\caption{Ablation study on fine-grained queries}
	\label{table:ab3}

			\begin{tabular}{cccc}
				\toprule
				\textbf{Query Type}                                                      & $\textbf{minFDE}_1$ & $\textbf{minADE}_5$ & $\textbf{minFDE}_5$ \\ \midrule
				\begin{tabular}[c]{@{}c@{}}Goal-based  queries\end{tabular}   & 7.08               & 1.20                & 2.30                \\
				\begin{tabular}[c]{@{}c@{}}Fine-grained queries\end{tabular} & \textbf{6.99}                & \textbf{1.17}               & \textbf{2.26}               \\ \bottomrule
				\end{tabular}
\end{table}

\subsubsection{Effect of the lane continuity}
We explicitly model lane segment continuity in multimodal trajectory prediction
 using GRU networks, ensuring temporal coherence between connected lane segments.
  In contrast, the baseline method mirrors to \cite{shi2024mtr++,zhou2023query} directly
aggregate all the nearby lane context without preserving the inherent 
continuity relationships of lane topology.
We demonstrate the performance of the two algorithms
 across distinct traffic scenarios similar to UniTraj \cite{feng2024unitraj} as follows:

\begin{itemize}
    \item \textbf{Straight driving}: Characterized by a heading difference
	 (end heading minus start heading) of less than 30$^\circ$
	  \textbf{and} lateral displacement 
	  of less than 5m, with no persistent unidirectional
	  drift trends.

    \item \textbf{Straight right/left}: Characterized by a heading difference (end heading minus start heading) of less than 30\(^\circ\) \textbf{but} lateral displacement exceeding 5m, with no persistent unidirectional turn trends.

    \item \textbf{Right/left turn}: Characterized by a heading difference (end heading minus start heading) exceeding 30\(^\circ\), regardless of lateral displacement magnitude or directional drift patterns.

\end{itemize}

As shown in Table \ref{tab:pred_errors}, our lane continuity-aware BLNet model demonstrates 
	consistent superiority over the baseline that neglects lane continuity 
	across, validating the effectiveness 
	of our proposed modeling strategy. In Straight Right/Left scenarios, 
	our model achieves a substantial 0.95 m (10.1 \%) 
	reduction in \(FDE_1\), underscoring robust modeling of complex
	 direction-changing behaviors. For Straight Driving scenarios, 
	 \(FDE_1\) and \(ADE_1\) are stably enhanced by 2.5--2.7\%,
	  reflecting superior precision for low-curvature, steady-state 
	  trajectories. While performing slightly worse than the 
	  baseline in Right/Left
	   Turn scenarios in \(FDE_5\) , aggregated results validate comprehensive 
	   superiority: \(FDE_1\) ($6.93 \rightarrow 6.67$) and \(ADE_1\) ($2.99 \rightarrow 2.89$). 
	   Collectively, our algorithm boosts prediction accuracy in mainstream 
	   scenarios, while maintaining robust overall
		 performance.

\begin{table}[!t]
\renewcommand{\arraystretch}{1.1}
\caption{Prediction Error Across Scenarios}
\label{tab:pred_errors}
\centering
\setlength{\tabcolsep}{2pt} 
\begin{tabular}{lcccccccc}
\toprule
\multirow{2}{*}{Scenario} & \multicolumn{2}{c}{FDE$_1$} & \multicolumn{2}{c}{ADE$_1$} & \multicolumn{2}{c}{FDE$_5$} & \multicolumn{2}{c}{ADE$_5$} \\
\cmidrule(lr){2-3} \cmidrule(lr){4-5} \cmidrule(lr){6-7} \cmidrule(lr){8-9}
& Base & BLNet & Base & BLNet & Base & BLNet & Base & BLNet \\
\midrule
Straight & 6.31 & \textbf{6.14} & 2.75 & \textbf{2.68} & 1.93 & \textbf{1.90} & 1.06 & \textbf{1.05} \\
Straight Right/Left & 9.45 & \textbf{8.50} & 3.95 & \textbf{3.63} & 3.30 & \textbf{3.14} & 1.57 & \textbf{1.54} \\
Right/Left Turn & 8.42 & \textbf{8.18} & 3.59 & \textbf{3.52} & \textbf{2.90} & 2.94 & \textbf{1.47} & \textbf{1.47} \\
\midrule
Overall & 6.93 & \textbf{6.67} & 2.99 & \textbf{2.89} & 2.21 & \textbf{2.18} & 1.17 & \textbf{1.16} \\
\bottomrule
\end{tabular}
\end{table}

\subsubsection{Effect of the Cross-Attention Mechanism}
To further investigate the effectiveness of different components in
 aggregating contextual information, we replace the cross-attention module 
 within the dual attention branch with alternative mechanisms. Specifically, 
 we select two mainstream context-aggregating modules widely adopted in 
 the trajectory prediction domain:

			\begin{itemize}
				
				\item Multi-context gating (MCG) \cite{varadarajan2022multipath++,li2023planning}: Multi-context gating (MCG) achieves
				 efficient fusion of multimodal information by stacking multiple Context
				  Gating (CG) blocks: it transforms set elements and context vectors 
				  via MLPs, fuses them through element-wise multiplication in a
				   permutation-invariant/equivariant manner, and then updates 
				   the context through pooling.
				\item Relative interaction block (RIB) \cite{kang2024ffinet,liang2020learning}: RIB models the associations between elements 
				by encoding explicit relative geometric features via MLPs, fuses these
				 relational features with element characteristics through concatenation or 
				 multiplication, and updates element representations by aggregating
				  context-aware interactive information, without relying on predefined 
				  graph structures.

			\end{itemize}

			To ensure efficiency and fairness in comparison, neither these variants
			 nor the baseline model underwent additional refinement. This experimental
			  design enables a focused evaluation of their intrinsic impact on the
			   backbone's overall performance. The comparative results are summarized
			    in Table \ref{table:ca}:

			\begin{table}[]
	\centering
	\caption{Ablation study on the attention block}
	\label{table:ca}
			\resizebox{\columnwidth}{!}{%
			\begin{tabular}{lcccc}
\toprule
Attention Type  & $ADE_1$ &$ FDE_1 $& $ADE_5$ &$ FDE_5$ \\
\midrule
MCG             & 3.04   & 7.10    & 1.21   & 2.36   \\
RIB             & 3.07   & 7.20    & 1.22   & 2.41   \\
  Cross Attention & \textbf{3.00} & \textbf{6.99} & \textbf{1.17} & \textbf{2.26} \\
\bottomrule
\end{tabular}
}
\end{table}

		Our cross-attention outperforms MCG and RIB across all metrics 
		(e.g., 3.3\% lower $ADE_5$ and 4.2\% lower $FDE_5$ 
		than MCG; 4.1\% lower $ADE_5$ and 6.2\% lower $FDE_5$ than RIB). 
		Unlike MCG's pooling-induced fine-grained loss and RIB's 
		reliance on explicit geometric features, cross-attention 
		enables adaptive weighting of critical multimodal cues, 
		confirming its superiority.

\subsection{Qualitative Analysis}

\subsubsection{Multi-scenario Visualization}

\begin{figure*}
			\centering
			\includegraphics[width=17cm]{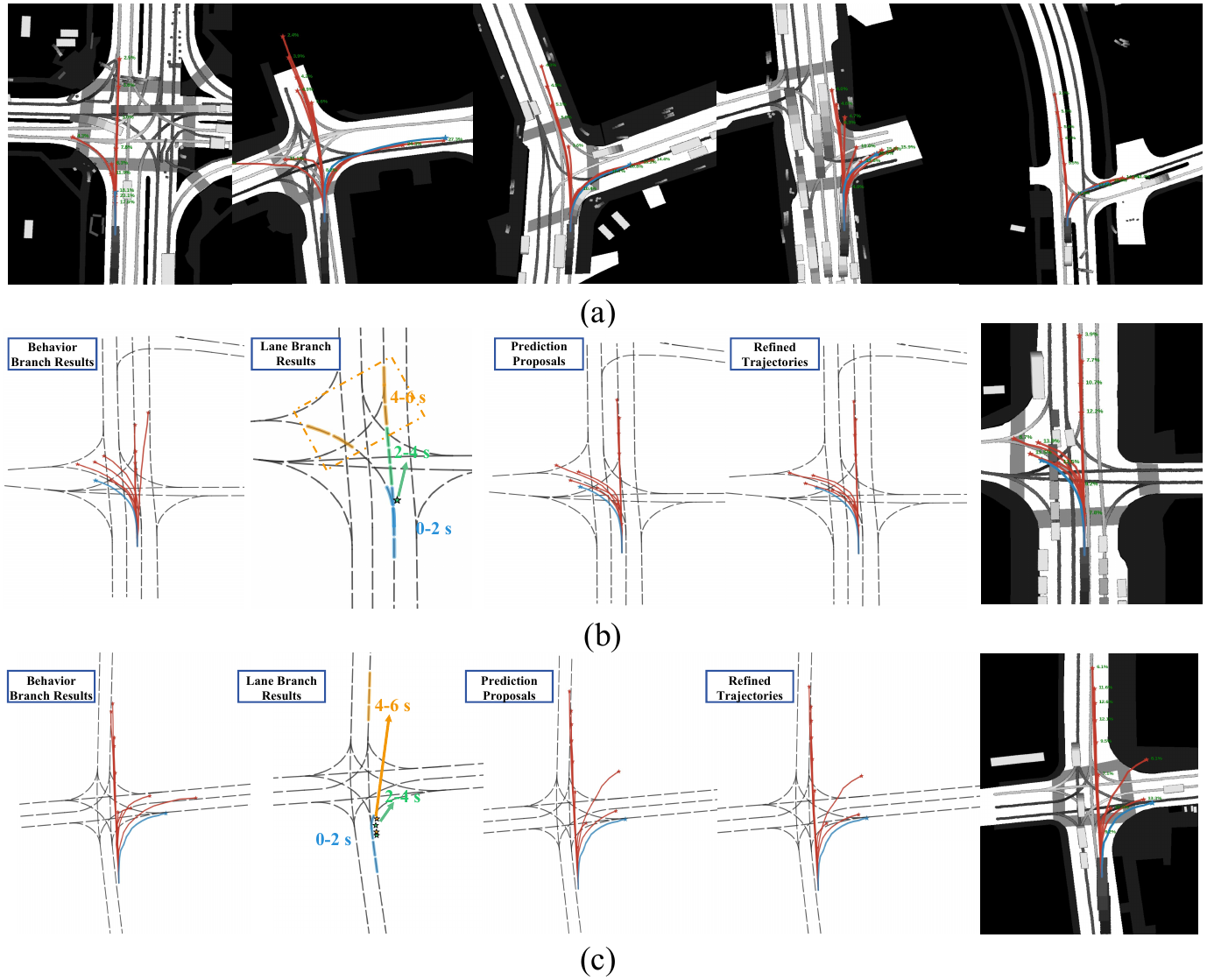}
\caption{Qualitative analysis results on the NuScenes validation set. The ground-truth future trajectory is shown in blue, while the predicted trajectories are shown in red. For lane prediction visualization, lane segments predicted for the 0--2~s, 2--4~s, and 4--6~s horizons are represented in blue, green, and yellow, respectively. If a lane segment is revisited by a later prediction interval, a star marker with the corresponding color is added to indicate the subsequent visitation. (a) Representative multimodal trajectory prediction results in diverse urban traffic scenarios. (b) Decomposition of the cascaded prediction process, illustrating how behavior-state hypotheses and lane constraints are jointly integrated to generate lane-consistent trajectory proposals and refined predictions. (c) A complementary case showing how the behavior-state branch compensates for imperfect lane guidance and helps recover the correct future motion through proposal generation and refinement.}
			\label{fig:vis1}
		\end{figure*}

To further illustrate the effectiveness of the proposed framework, we provide qualitative visualization results  as shown in Fig.~\ref{fig:vis1}. In all visualizations, the ground-truth future trajectory is represented by the blue curve, while the predicted trajectories are shown in red. For the lane attention branch visualization, directly displaying the predicted lane segments for every future timestamp would result in severe visual clutter. Therefore, for clarity, the predicted lane segments associated with different future horizons are grouped and color-coded according to their corresponding prediction intervals. Specifically, blue lane segments denote the lane candidates predicted for the 0--2 s horizon, green lane segments correspond to the 2--4 s horizon, and yellow lane segments indicate the 4--6 s horizon. When a lane segment is revisited by a later prediction interval, a star marker with the corresponding color is added beside the lane segment to indicate the subsequent traversal, thereby visualizing the temporal evolution of lane predictions throughout the forecasting horizon.

Fig.~\ref{fig:vis1}(a) presents representative multimodal prediction results across diverse urban traffic scenarios. The predicted trajectories generally cover ground truth future motion patterns and remain consistent with the surrounding lane geometry, demonstrating that the proposed model can capture both motion diversity and lane-constrained trajectory evolution.

Fig.~\ref{fig:vis1}(b) further decomposes the cascaded prediction process in a representative scenario. The behavior-state branch is able to generate diverse future motion hypotheses. Nevertheless, several trajectory modes are observed to drift toward unreasonable lane segments because no explicit lane topology constraints are imposed at this stage. The lane attention branch compensates for this limitation by predicting the lane segments likely to be occupied at different future timestamps and providing fine-grained lane guidance. As a result, the generated proposal becomes more consistent with the actual road structure. Afterward, the refinement module further adjusts the proposal using future motion features and lane continuity constraints, producing a final trajectory that more closely matches the ground-truth motion.

Fig.~\ref{fig:vis1}(c) shows another representative case where the lane branch alone provides imperfect lane guidance. Specifically, the most probable lane segments predicted by the lane branch do not fully cover the actual future path of the target agent. Nevertheless, the behavior-state branch preserves a plausible motion mode that complements the missing lane guidance and enables the first-stage proposal to remain close to the ground-truth trajectory. After the refinement stage, the final prediction is further corrected and becomes more consistent with the actual future motion.

These visualizations demonstrate that the proposed dual-stream design does not simply rely on either behavioral intention or lane topology alone. Instead, the behavior-state branch and lane attention branch provide complementary guidance, while the refinement module further improves local trajectory consistency. This cascaded process enables BLNet to generate more accurate and physically reasonable predictions under different types of uncertainty.

\subsubsection{Failure Case Analysis}
Although BLNet generally produces accurate and lane-consistent predictions, prediction errors may still occur in highly uncertain scenarios.

\begin{figure*}
			\centering
			\includegraphics[width=17cm]{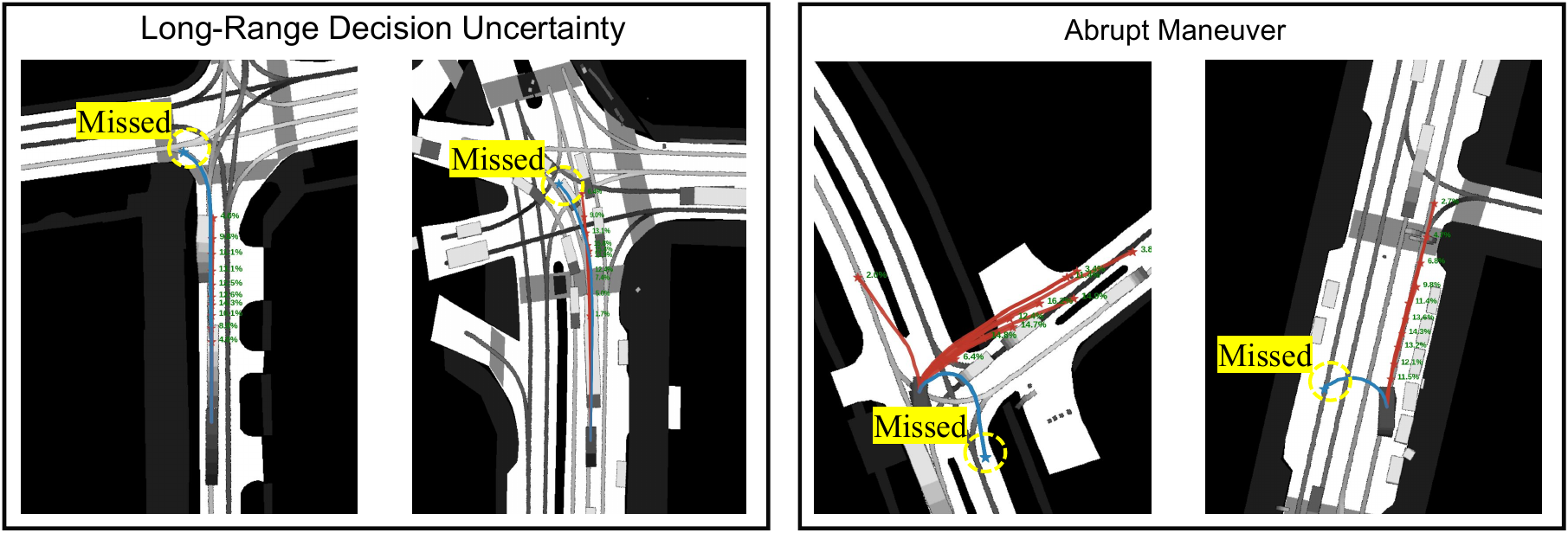}
\caption{Representative failure cases on the NuScenes validation set. The ground-truth future trajectory is shown in blue, while the predicted trajectories are shown in red. Left: a failure case occurring before a decision point, where the target vehicle remains relatively far from the upcoming maneuver and limited motion evidence leads to increased uncertainty in future intention inference. Right: a failure case involving an abrupt maneuver, where a sudden large-direction change causes the predicted trajectories to deviate from the ground truth. These examples highlight the challenges of early decision ambiguity and rare long-tail driving behaviors in trajectory prediction.}
			\label{fig:vis2}
		\end{figure*}

The left example of Fig. \ref{fig:vis2} presents a failure case that occurs far from a decision point. At this stage, the target vehicle remains relatively distant from the upcoming intersection, and the observed motion history provides insufficient cues regarding the final maneuver intention. Consequently, multiple future behaviors remain plausible, making it difficult for the model to accurately identify the correct future maneuver. Although the generated trajectories remain physically feasible, the dominant prediction deviates from the ground-truth future path. This example highlights the intrinsic difficulty of long-horizon prediction before decision regions, where future intentions have not yet been clearly expressed through observable motion patterns.

The right example of Fig. \ref{fig:vis2} illustrates a failure case involving an abrupt maneuver. In this scenario, the target vehicle performs a sudden large-direction change that is weakly represented in the training data. As a result, both the lane topology cues and historical motion patterns provide limited evidence for the future maneuver, causing the generated trajectory hypotheses to deviate from the ground truth. This example demonstrates the challenges posed by long-tail driving behaviors and rare maneuver patterns.

Overall, these failure cases indicate that the primary sources of prediction errors arise from early decision ambiguity and rare abrupt maneuvers. Addressing such challenges may require stronger intention reasoning capabilities and improved modeling of long-tail driving behaviors, which will be investigated in future work.

\subsection{Discussion on Architectural Novelty}
This section further discusses the architectural novelty 
of BLNet from the perspective of fine-grained trajectory 
prediction. Rather than relying on any individual architectural 
component, the proposed framework is built upon a unified formulation that jointly models behavioral evolution and lane topology constraints throughout the prediction horizon.

 \begin{itemize}
	\item Fine-Grained Dual-Stream Token Design

	BLNet adopts a target-agent-centric dual-stream 
	architecture with timestamp-level behavioral state 
	tokens and lane topology tokens. Each learnable token branch 
	is supervised
	 by dedicated auxiliary objectives, enabling behavioral 
	 evolution and lane topology constraints to be modeled explicitly
	  and optimized simultaneously. Specifically, the behavioral-state
	   branch focuses on capturing the dynamic evolution of 
	   driving intentions throughout the prediction horizon, 
	   while the lane branch characterizes the corresponding lane topology 
	   constraints that may influence future motion.
Consequently, future trajectories are generated through the
joint modeling of these two complementary factors, enabling
future motion to be characterized through explicit behavioral
evolution and lane topology constraints rather than a single
motion hypothesis.
	\item Point-Level Trajectory Refinement with Spatiotemporal Continuity

	 Instead of directly aggregating all 
	 nearby contextual information into queries,
	  it fuses sequential lane
	  continuity features and future motion 
	  features to achieve point-level trajectory refinement---with 
	  tangible performance gains validated by our ablation 
	  study (see Table \ref{tab:pred_errors}). This design 
	  enables the predicted trajectories to better preserve local motion consistency and lane-following continuity
	   and effectively reduces prediction errors 
	  in multiple driving scenarios.

\item Joint Supervision of Behavioral Evolution and Lane Constraints

To support the proposed fine-grained trajectory prediction formulation, BLNet introduces dedicated auxiliary supervision for the behavior-state and lane query branches. By jointly optimizing behavioral evolution and lane topology constraints, the proposed supervision strategy encourages consistent motion-state representations throughout the prediction horizon, thereby improving trajectory generation quality.

 \end{itemize}

\section{Conclusion and Future Work}

In this study, we propose a novel trajectory prediction framework integrating fine-grained lane constraints and behavioral evolution via a dual-stream attention architecture. The model contains two complementary branches: one captures lane topology constraints, and the other learns latent behavioral patterns. We adopt a two-stage decoding strategy to generate trajectory proposals and refine trajectories with lane continuity and motion features. Experiments on nuScenes and Argoverse verify our framework's effectiveness, with ablation studies confirming each module's contribution.

Future work will further explore the quantitative correlation between the temporal distance to decision points and prediction performance. Potential optimization schemes will be studied to enhance model forecasting capability for scenarios where observations are obtained long before decision points with ambiguous driving intentions. 

In addition, our lane-continuity refinement has two clear limitations in intersections, merging areas, lane-changing and map-uncertain scenarios. It only locally adjusts trajectories within nearby lane structures and cannot fix severely biased initial proposals from abrupt maneuvers. Moreover, it depends on accurate HD maps; ambiguous lane branches or defective map data will weaken its refinement effect. We will develop more robust refinement strategies jointly modeling lane ambiguity, unreliable map information and interactive behaviors to strengthen prediction generalization in these challenging scenarios.

Furthermore, the current learnable behavior and lane tokens rely purely on data-driven latent learning with limited interpretability, which may degrade performance on rare and long-tail traffic cases. Future work will introduce semantics-guided interpretable tokens to enhance robustness and generalization.

\bibliographystyle{IEEEtran} 
\bibliography{IEEEabrv,myref}

\begin{IEEEbiography}[{\includegraphics[width=1in,height=1.25in,clip,keepaspectratio]{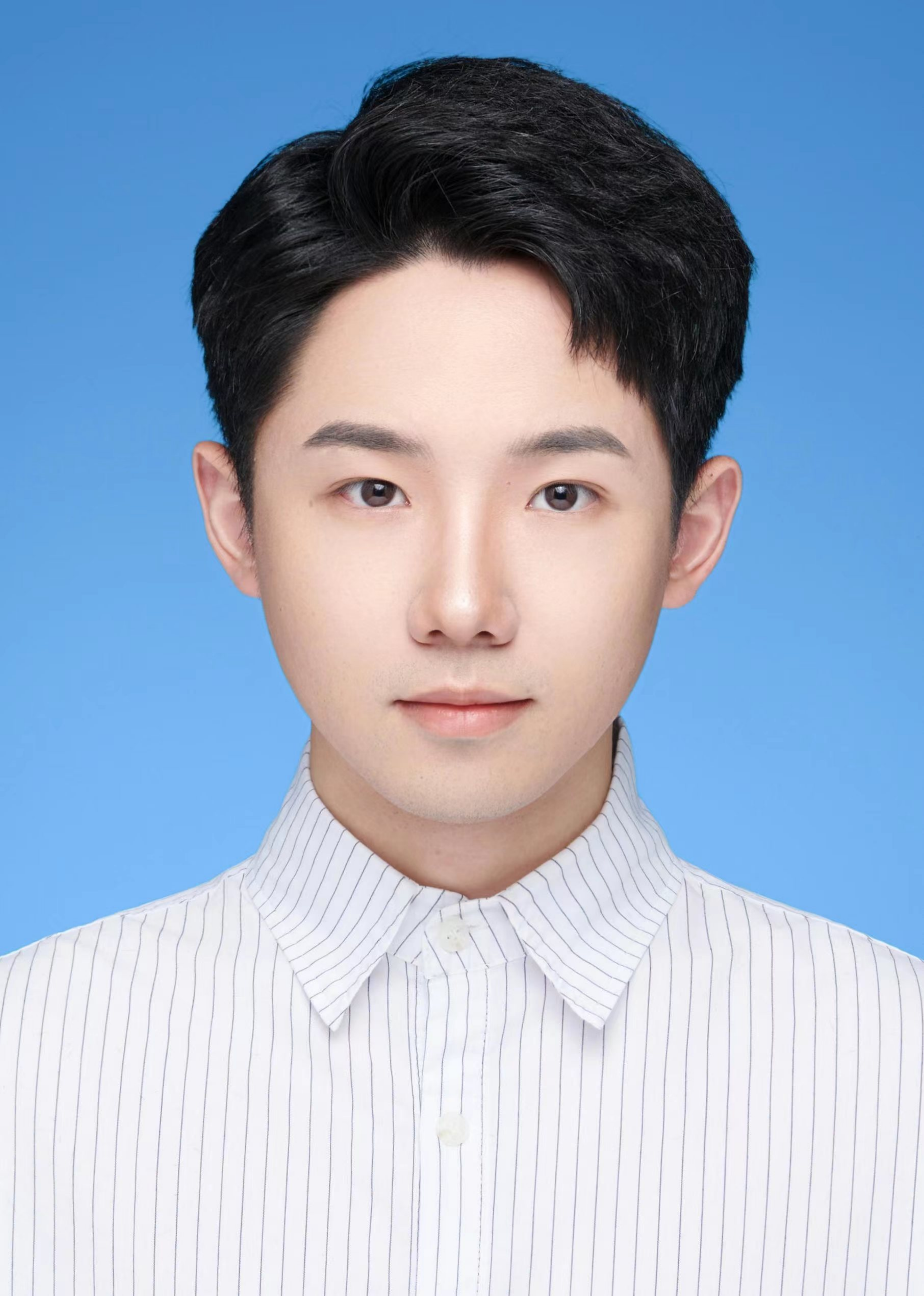}}]
	{Wenyi Xiong} received his B.E. degree in mechanical and electrical engineering from Central South University, Changsha, China, in 2021. He is currently working toward the Ph.D. degree in the College of Mechanical Engineering, Zhejiang University, Hangzhou, China.
	
	His research interests include vehicle motion prediction, trajectory planning, and deep learning.
\end{IEEEbiography}
\begin{IEEEbiography}[{\includegraphics[width=1in,height=1.25in,clip,keepaspectratio]{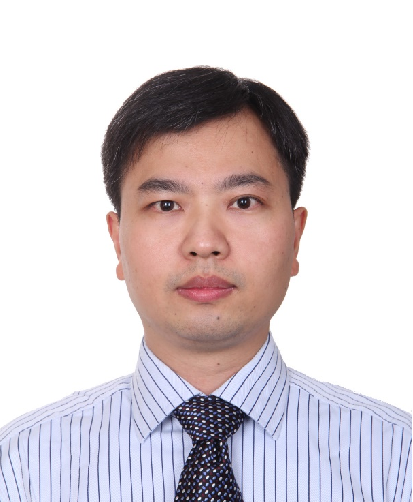}}]
	{Jian Chen} (Senior Member, IEEE) received the B.E. and M.E. degrees from Zhejiang University, Hangzhou, China, in 1998 and 2001, respectively, and the Ph.D. degree in electrical engineering from Clemson University, Clemson, SC, USA, in 2005.
	He was a Research Fellow with the University of Michigan, Ann Arbor, MI, USA, from 2006 to 2008, where he was involved in fuel cell modeling and control. From 2013 to 2024, he was a professor in the College of Control Science and Engineering, Zhejiang University. Currently, he is a professor in the School of Automation and Intelligent Manufacturing, Southern University of Science and Technology, Shenzhen, China. His research interests include modeling and control of fuel cell systems, visual servo techniques, battery management systems, and applied nonlinear control.
\end{IEEEbiography}
\begin{IEEEbiography}[{\includegraphics[width=1in,height=1.25in,clip,keepaspectratio]{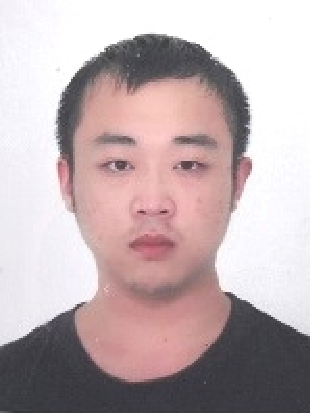}}]
	{Qi Ziheng} received his Bachelor's degree in 2019 and Master's degree in 2022, both from the College of Control Science and Engineering, Zhejiang University.
	
His research interests include vehicle dynamics, path planning, and tracking of autonomous vehicles.
\end{IEEEbiography}
\end{document}